%% file: main_arxiv.tex
\documentclass[10pt,twocolumn,letterpaper]{article}

\usepackage[pagenumbers]{cvpr}       

\input{preamble}

\definecolor{cvprblue}{rgb}{0.21,0.49,0.74}
\definecolor{arxivurlrose}{HTML}{BE3156}
\usepackage[breaklinks,colorlinks,allcolors=cvprblue]{hyperref}

\def\confName{CVPR}
\def\confYear{2027}

\title{\textcolor{navworldred}{RecallWorld}: Episodic Memory for Persistent Embodied Video World Models}

\author{\parbox{\textwidth}{\centering\normalsize
Sarosij Bose\textsuperscript{1,2}\quad
Gitika Karumuri\textsuperscript{1}\quad
Chenqi Li\textsuperscript{1,3}\quad
Allen Tu\textsuperscript{1,4}\quad
Srikumar Sastry\textsuperscript{1,5}\\
Mellon Zhang\textsuperscript{1,6}\quad
Avinash Ravichandran\textsuperscript{1}\quad
Alexander Schwing\textsuperscript{1,7}\quad
Konstantinos Karydis\textsuperscript{2}\quad
Amit K. Roy Chowdhury\textsuperscript{2}\\[3pt]
{\small \textsuperscript{1}Amazon\quad
\textsuperscript{2}University of California, Riverside\quad
\textsuperscript{3}University of Oxford\quad
\textsuperscript{4}University of Maryland, College Park\\
\textsuperscript{5}Washington University in St. Louis\quad
\textsuperscript{6}Georgia Institute of Technology\quad
\textsuperscript{7}University of Illinois Urbana-Champaign}\\[3pt]
\href{https://recallworld.github.io/}{\textcolor{arxivurlrose}{\texttt{https://recallworld.github.io/}}}
}}

\begin{document}

\raggedbottom
\enableteaser
\maketitle
\input{sections/00_abstract}
\setcounter{dbltopnumber}{1}
\input{sections/arxiv/01_introduction}

\input{figures/method/figure}
\input{sections/02_related_work}
\input{sections/arxiv/03_method}
\input{sections/arxiv/04_experiments}
\input{sections/arxiv/05_results}
\makeatletter
\let\recallworldclosingsection\cvprsection
\def\cvprsection{\@startsection{section}{1}{\z@}
  {-4pt plus -1pt minus -1pt}{3pt}{\large\bf}}
\input{sections/07_conclusion}
\input{sections/06_limitations}
\let\cvprsection\recallworldclosingsection
\makeatother

\input{sections/arxiv/08_acknowledgements}
{\small
\bibliographystyle{ieeenat_fullname}
\bibliography{main}
}
\end{document}

%% file: preamble.tex
\usepackage[T1]{fontenc}
\usepackage{colortbl}
\usepackage{tikz}
\usepackage{tabularx}
\usepackage{multirow}
\usepackage{pifont}

\newcommand{\methodname}{RecallWorld}
\newcommand{\navworldperspective}{\methodname{}}
\newcommand{\navworldfisheye}{\methodname{}-PF}
\newcommand{\queryset}{\mathcal Q}
\newcommand{\historybank}{\mathcal M}
\newcommand{\scenelet}{\mathcal S}
\newcommand{\keyframes}{\mathcal J}
\newcommand{\selectedset}{\mathcal G}
\newcommand{\lossname}{\mathcal L}

\makeatletter
\newcommand{\relatedparagraph}[1]{%
  \@startsection{paragraph}{4}{\z@}%
    {4pt plus 1pt minus 1pt}{-1em}%
    {\normalfont\normalsize\bfseries}{#1}}
\makeatother

\definecolor{navworldred}{HTML}{941F30}
\definecolor{navworldrow}{HTML}{FAEFE8}
\colorlet{navworldcomparisonrow}{navworldrow}
\definecolor{navworldcontrol}{HTML}{95630D}
\colorlet{capabilitygreen}{green!60!black}
\colorlet{capabilityred}{red!75!black}
\newcommand{\capabilityyes}{\textcolor{capabilitygreen}{\ding{51}}}
\newcommand{\capabilityno}{\textcolor{capabilityred}{\ding{55}}}
\let\navworldplainname\methodname
\newcommand{\navworldtablestyle}{%
  \renewcommand{\methodname}{%
    \textcolor{navworldred}{\textbf{\navworldplainname}}}%
  \renewcommand{\navworldperspective}{%
    \textcolor{navworldred}{\textbf{\navworldplainname}}}%
  \renewcommand{\navworldfisheye}{%
    \textcolor{navworldred}{\textbf{\navworldplainname-PF}}}}

\graphicspath{{figures/}{figures/suppl/}}

\AtBeginEnvironment{equation}{%
  \setlength{\abovedisplayskip}{8pt plus 2pt minus 2pt}%
  \setlength{\belowdisplayskip}{8pt plus 2pt minus 2pt}%
  \setlength{\abovedisplayshortskip}{3pt plus 1pt}%
  \setlength{\belowdisplayshortskip}{6pt plus 2pt minus 2pt}%
}

\definecolor{saromaroon}{RGB}{128,0,32}

\newcommand{\upmetric}{\ensuremath{\uparrow}}
\newcommand{\downmetric}{\ensuremath{\downarrow}}

\makeatletter
\newsavebox{\navworldcategorybox}
\newcommand{\navworldtablepart}[2]{%
  \sbox{\navworldcategorybox}{\scriptsize\bfseries #2}%
  \smash{\raisebox{\dimexpr
    (2\ht\@arstrutbox-#1\ht\@arstrutbox-#1\dp\@arstrutbox
    -\ht\navworldcategorybox+\dp\navworldcategorybox)/2\relax}{%
    \usebox{\navworldcategorybox}}}}
\newcommand{\navworldcategory}[3]{%
  \sbox{\navworldcategorybox}{%
    \rotatebox{90}{\scriptsize\bfseries
      \textcolor{#2}{\shortstack{#3}}}}%
  \smash{\raisebox{\dimexpr
    (2\ht\@arstrutbox-#1\ht\@arstrutbox-#1\dp\@arstrutbox
    -\ht\navworldcategorybox+\dp\navworldcategorybox)/2\relax}{%
    \usebox{\navworldcategorybox}}}}
\newcommand{\enableteaser}{%
  \apptocmd{\@maketitle}{\input{figures/teaser/figure}}{}{%
    \PackageError{navworld}{Could not attach teaser to title block}{}%
  }%
}
\makeatother

%% file: figures/teaser/figure.tex
\begin{minipage}{\textwidth}
  \centering
  \input{figures/teaser/layout}
  \captionsetup{hypcap=false,skip=2pt}
  \captionof{figure}{%
    \textbf{\methodname{}} consolidates perspective and fisheye views into episodic
    Scenelets that condition the world model to generate faithful revisits along
    the target camera trajectory. Retrieving relevant Scenelets on revisits
    preserves scene structure and appearance that prior methods often lose
    over long-horizon generation. When the camera
    revisits a pose in GrandTour~\cite{tuna2026grandtour}, prior methods~\cite{zhu2026sanawm,ren2025gen3c,zhou2025seva,zhao2025spatia} alter
    scene identity, whereas episodic memory preserves the observed railway
    structure and appearance.
  }
  \label{fig:teaser}
\end{minipage}
\par\vspace{5pt}

%% file: figures/teaser/layout.tex
\begingroup%
\begin{tikzpicture}[baseline=(current bounding box.south),
  x=\textwidth/5220,y=-\textwidth/5220]
\path[use as bounding box] (0,0) rectangle (5220,1443);
\node[anchor=north west,inner sep=0pt,outer sep=0pt] at (0,0)
 {\includegraphics[width=\textwidth]{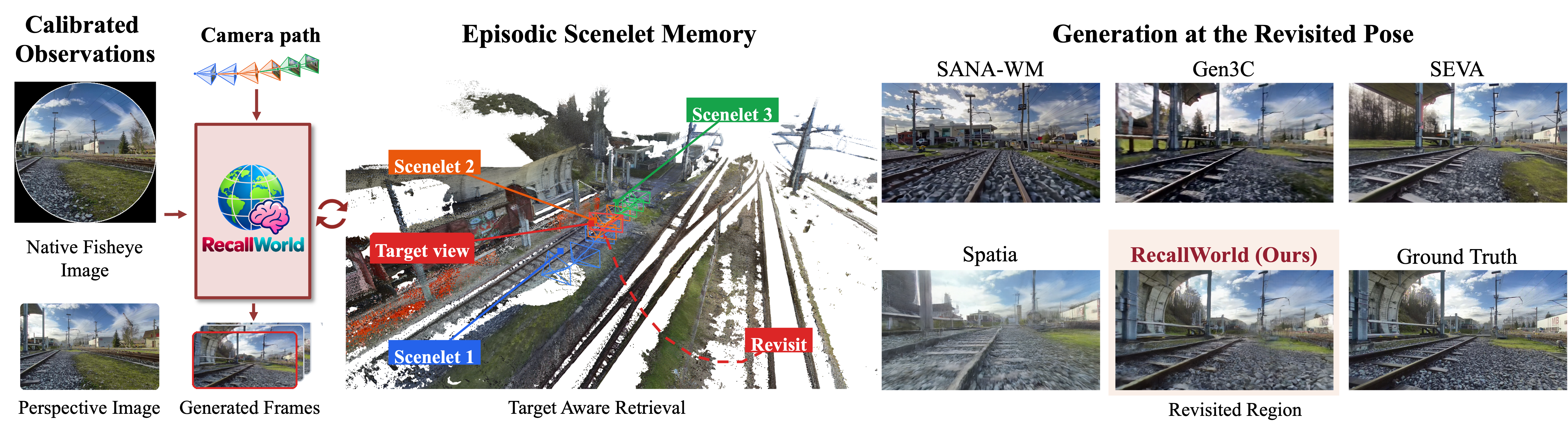}};
\fill[white] (20,25) rectangle (565,235);
\node[anchor=center,inner sep=0pt,outer sep=0pt,
  font=\rmfamily\fontsize{7.5}{8.5}\selectfont\bfseries]
  at (285,123.5) {Source Views};
\fill[white] (660,80) rectangle (1080,163);
\node[anchor=center,inner sep=0pt,outer sep=0pt,
  font=\rmfamily\fontsize{7.5}{8.5}\selectfont\bfseries]
  at (869,123.5) {Camera Path};
\definecolor{recallteaserorange}{RGB}{234,88,12}
\definecolor{recallteaserblue}{RGB}{37,99,235}
\definecolor{recallteasergreen}{RGB}{22,163,74}
\definecolor{recallteaserred}{RGB}{220,38,38}
\fill[recallteaserorange] (1294,500) rectangle (1600,581);
\fill[recallteaserblue] (1294,1134) rectangle (1600,1215);
\fill[recallteasergreen] (2286,326) rectangle (2592,407);
\fill[recallteaserred] (2481,1093) rectangle (2705,1174);
\tikzset{recallteaserlabel/.style={
  anchor=center,inner sep=0pt,outer sep=0pt,text=white,
  font=\rmfamily\fontsize{6.5}{7.5}\selectfont\bfseries}}
\node[recallteaserlabel] at (1447,540.5) {Scenelet 2};
\node[recallteaserlabel] at (1447,1174.5) {Scenelet 1};
\node[recallteaserlabel] at (2439,366.5) {Scenelet 3};
\node[recallteaserlabel] at (2593,1133.5) {Revisit};
\fill[navworldrow] (3755,816) rectangle (4393,895);
\node[anchor=center,inner sep=0pt,outer sep=0pt,text=navworldred,
  font=\rmfamily\fontsize{7.5}{8.5}\selectfont\bfseries]
  at (4074,854.5) {\methodname{} (OURS)};
\end{tikzpicture}%
\endgroup%

%% file: sections/00_abstract.tex
\begin{abstract}
\looseness=-1
Video world models generate plausible novel views but often lack persistent memory,
failing to preserve a place's geometry and appearance when an agent revisits a scene.
A central challenge is how previous views should be effectively used for creating
memory. Reconstructing them independently underutilizes geometric constraints
between nearby views, while repeatedly reconstructing a growing history can
propagate later prediction errors into earlier geometry. We introduce RecallWorld,
a persistent embodied video world model that jointly reconstructs short multi-view
episodes into Scenelets while keeping geometry independent across episodes.
Starting from a single source image, camera-aware ray lifting places perspective
and native fisheye images in a common world frame, retaining wide-field content for
perspective generation. Each generated segment extends the Scenelet memory
and conditions subsequent generation. Coverage-based retrieval selects
complementary Scenelets from the full history, while target-view visibility is
resolved geometrically before appearance fusion, providing consistent memory
conditioning for guiding subsequent generation. Across four datasets, RecallWorld
improves revisit PSNR over the strongest evaluated baseline by 2.6 dB on partial
revisits and 3.9 dB on full cycles, demonstrating substantially stronger preservation
of scene identity across exploration and return.
\par
\end{abstract}

%% file: sections/arxiv/01_introduction.tex
\section{Introduction}
\label{sec:introduction}
\suppressfloats[t]

Video world models generate future visual observations conditioned on actions
or camera motion, supporting robotic interaction and
manipulation~\cite{yang2023unisim,zhou2024robodreamer,wu2024ivideogpt},
autonomous driving~\cite{gao2024vista,russell2025gaia2,zhao2024drivedreamer2},
and synthetic training-data
generation~\cite{nvidia2025cosmos,jang2025dreamgen,ye2026robodream}. However, for embodied
exploration, plausible next frames are not enough: the model must preserve a
place's geometry and appearance after its observations leave the recent context as an agent may later approach the same location from a 
new direction. Persistent video generation therefore requires long-term access to
earlier observations without reconstructing or conditioning on the full history.

No prior video world model combines these capabilities. Camera-controlled
generators produce locally coherent video
~\cite{bar2024nwm,wan2025wan,gao2026lingbotworld,he2025cameractrlii,zhu2026sanawm}
but do not guarantee consistent revisits. Memory-based methods retrieve earlier
views or project stored 3D content
~\cite{xiao2026worldmem,ren2025gen3c,li2025vmem,wu2025spmem}, but
\emph{leave the granularity at which memory should be consolidated unresolved}.
Independent per-view reconstruction discards
constraints among nearby observations, whereas repeatedly reconstructing the
growing history lets later
generated content revise earlier geometry. We design persistent memory by
reconstructing nearby views jointly as \emph{Scenelets}, keeping episode geometry independent,
and retrieving target-relevant content within a fixed conditioning budget.

We present \textbf{\methodname{}}, a persistent embodied video world model with
episodic Scenelet memory. Firstly, camera-aware ray lifting unifies perspective and native fisheye
observations in a common world frame~\cite{piccinelli2025unik3d}, retaining
surfaces outside the perspective image used for generation. Next, each completed segment forms a Scenelet:
overlapping views are reconstructed jointly, and geometry remains independent
across episodes. Then, coverage-based retrieval utilizes complementary Scenelets under a fixed conditioning budget,
and target-view geometry resolves visibility before appearance fusion. This design separates local reconstruction from long-term retention while keeping
earlier Scenelets accessible. Figure~\ref{fig:method} illustrates our
pipeline.

We evaluate \methodname{} on real robot-navigation trajectories from
GrandTour~\cite{tuna2026grandtour} and M2DGR~\cite{yin2021m2dgr}, synthetic
trajectories from TartanGround~\cite{patel2025tartanground}, and real scene
captures from DL3DV~\cite{ling2023dl3dv}. On exploration-and-return trajectories
starting from the same source image, \methodname{}'s generated revisits preserve
the structure and appearance of previously observed places, whereas
camera-controlled and global-memory baselines alter scene content. Our ablation
studies show that Scenelets improve revisit fidelity over independent views as
memory and reduce peak reconstruction memory relative to global memory. Natively using
fisheye images further expands the surface coverage retained in memory.
We also measure camera adherence and video coherence throughout each rollout. Together, these
results show that \methodname{} preserves place identity across exploration
and partial or full return cycles.

Our main contributions are:
\begin{itemize}[itemsep=2pt,topsep=3pt,parsep=0pt]
\item \textbf{Episodic Scenelet memory.}
We show that how observations are grouped into memory is important for
persistent video generation. We introduce episodic Scenelets, which jointly
reconstruct nearby views while keeping geometry independent across different
episodes.

\item \textbf{Persistent Retrieval and visibility-aware conditioning.}
We retrieve Scenelets from the full history under a fixed memory budget and
use their geometry to determine which surfaces are visible in the target view
before combining appearance for generation.

\widowpenalty=10000
\item \textbf{Camera general persistent representation.}
We use camera-aware ray lifting to retain perspective and native fisheye
observations in a common spatial representation, preserving wide-field surfaces
for later perspective generation without changing the Scenelet pipeline.
\end{itemize}

\input{tables/01_memory_design}

%% file: tables/01_memory_design.tex
\begin{table}[t]
\centering\small
\navworldtablestyle
\setlength{\tabcolsep}{2.25pt}
\renewcommand{\arraystretch}{1.12}
\caption{\textbf{Memory and conditioning in video world models.}\\
\methodname{} combines persistent 3D memory, navigation video generation, and
local 3D units with independent multi-view reconstructions.
\navworldfisheye{} extends this design by storing perspective and native
fisheye observations in a unified memory.}
\label{tab:memory-design}
\begin{tabularx}{\linewidth}{@{}l
 >{\hsize=.85\hsize\linewidth=\hsize\centering\arraybackslash}X
 >{\hsize=1.25\hsize\linewidth=\hsize\centering\arraybackslash}X
 >{\hsize=.95\hsize\linewidth=\hsize\centering\arraybackslash}X
 >{\hsize=.95\hsize\linewidth=\hsize\centering\arraybackslash}X@{}}
\toprule
\textbf{Method}
 & \shortstack{\textbf{3D}\\\textbf{Memory}}
 & \shortstack{\textbf{Navigation}\\\textbf{Video Gen.}}
 & \shortstack{\textbf{Local 3D}\\\textbf{Units}}
 & \shortstack{\textbf{Unified}\\\textbf{Memory}} \\
\midrule
SANA-WM~\cite{zhu2026sanawm}
 & \capabilityno & \capabilityno & \capabilityno & \capabilityno \\
NWM~\cite{bar2024nwm}
 & \capabilityno & \capabilityyes & \capabilityno & \capabilityno \\
\midrule
Gen3C~\cite{ren2025gen3c}
 & \capabilityyes & \capabilityno & \capabilityno & \capabilityno \\
VMem~\cite{li2025vmem}
 & \capabilityyes & \capabilityno & \capabilityno & \capabilityno \\
SPMem~\cite{wu2025spmem}
 & \capabilityyes & \capabilityno & \capabilityno & \capabilityno \\
Spatia~\cite{zhao2025spatia}
 & \capabilityyes & \capabilityno & \capabilityno & \capabilityno \\
Lyra 2.0~\cite{shen2026lyra2}
 & \capabilityyes & \capabilityno & \capabilityno & \capabilityno \\
\midrule
\rowcolor{navworldcomparisonrow}
\methodname
 & \capabilityyes & \capabilityyes & \capabilityyes
 & \capabilityyes$_{\color{navworldred}\mathbf{PF}}$ \\
\bottomrule
\end{tabularx}
\end{table}

%% file: figures/method/figure.tex
\begin{figure*}[t]
  \centering
  \includegraphics[width=\textwidth]{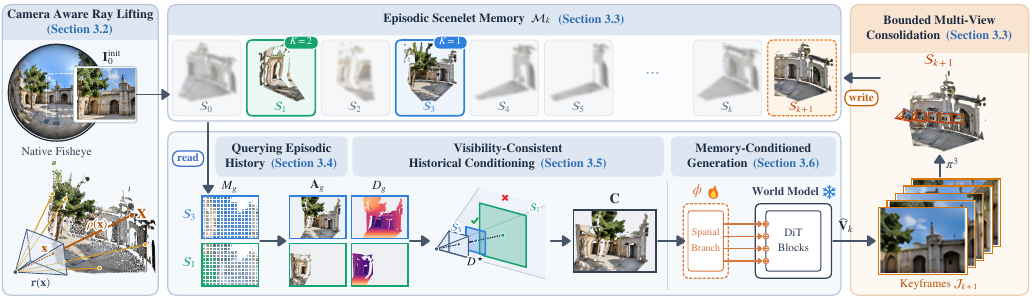}
  \captionsetup{skip=0pt}
  \caption{\textbf{Overview of \methodname{}.}
  Ray lifting unifies perspective and native fisheye inputs in 3D;
  consolidation forms episodic Scenelet memory $\historybank_k$.
  Complementary retrieval, target-view projection and visibility resolution
  form $\mathbf C_k$, which conditions
  \mbox{SANA-WM}~\cite{zhu2026sanawm} via a spatial-conditioning branch.
    Generated keyframes are jointly reconstructed into $\scenelet_{k+1}$
    for continued memory-aware generation.}
  \label{fig:method}
  \vspace{-4pt}
\end{figure*}

%% file: sections/02_related_work.tex

\section{Related Work}
\label{sec:related-work}

\noindent\textbf{Camera Conditioned Video Generation.}
Video diffusion models synthesize temporally coherent
sequences~\cite{ho2022videodiffusion,blattmann2023stablevideo}.
Video world models predict future observations from actions or camera motion.
Action-conditioned models support interactive
navigation~\cite{bruce2024genie,wu2024ivideogpt,bar2024nwm}, while
pose-conditioned diffusion and explicit 3D guidance enable broader camera
control~\cite{he2024cameractrl,he2025cameractrlii,cao2025uni3c}.
SANA-WM generates minute-scale videos along continuous camera
trajectories~\cite{zhu2026sanawm}, SEVA synthesizes image-conditioned novel
views~\cite{zhou2025seva}, and GEN3C uses a 3D cache for world-consistent
camera control~\cite{ren2025gen3c}.
Camera control, however, does not ensure a consistent revisit once a place
leaves the model's recent context.

\noindent\textbf{Feed Forward Reconstruction.}
Beyond camera control, feed-forward models recover geometry and
correspondences from images~\cite{wang2024dust3r,leroy2024mast3r},
while generative reconstruction jointly models geometry and appearance to ground
synthesized views in observed
structure~\cite{szymanowicz2025bolt3d,celen2026genrec,zhang2026worldstereo}.
Most generative models assume pinhole imagery, and rectifying wide field-of-view
inputs discards peripheral coverage~\cite{duan2026fisheye3r}.
UniK3D instead estimates rays and radial range for perspective, fisheye, and
panoramic cameras~\cite{piccinelli2025unik3d}.
Fisheye3R and Cam3R generalize feed-forward
reconstruction~\cite{duan2026fisheye3r,guruprasad2026cam3r}, while ray-based
generators extend camera control across lens
geometries~\cite{zhang2025unified,jin2026crepe,hu2026zytworld}.
\methodname{} uses $\pi^3\mathrm{X}$~\cite{wang2025pi3} for joint multi-view reconstruction,
Depth Anything~3~\cite{lin2025da3} for perspective initialization, and UniK3D~\cite{piccinelli2025unik3d}
for native fisheye initialization. Together, these components provide
camera-aware geometry for episodic memory.

\noindent\textbf{Memory Aware Generation.}
For persistence over long horizons, video models retain spatial evidence as retrieved images indexed
by pose, time or field-of-view overlap~\cite{xiao2026worldmem,yu2025contextmemory},
projected 3D caches or RGB-D features~\cite{zhou2025persistentembodied,he2026ppmem},
and evolving global or streaming
maps~\cite{chen2026anyrecon,yu2026worldcrafter,jin2026zipmap,li2026closer}.
Memory organization then governs which observations share a reconstruction and
remain available for future queries.
VMem indexes earlier views with surfels~\cite{li2025vmem}, SPMem combines
geometric support with recent and episodic images~\cite{wu2025spmem}, and
Spatia maintains an updatable point-cloud memory~\cite{zhao2025spatia}.
AnchorWeave retrieves complementary frame-associated geometry by marginal
coverage~\cite{wang2026anchorweave}.
\methodname{} follows this retrieval family while changing the persistent
memory unit: nearby views are jointly reconstructed before retention,
whereas separate episodes retain independent geometry and are reconciled
after target-view projection.
Table~\ref{tab:memory-design} compares these components.

%% file: sections/arxiv/03_method.tex
\section{\methodname{}}
\label{sec:method}

A persistent video world model must preserve observed geometry and appearance
after a place leaves its recent context. \textbf{\methodname{}} generates video
autoregressively and consolidates each completed segment into an independent
\textbf{Scenelet}. Nearby keyframes share a joint reconstruction, while stored Scenelets
remain fixed, preventing later prediction errors from rewriting earlier
geometry. Camera-aware ray lifting places perspective and fisheye observations in
a common world-frame representation. Before generating each segment,
\methodname{} retrieves a complementary set of Scenelets covering the target
views, resolves visibility among their projected surfaces, and conditions the
generator on the resulting appearance. This separates local reconstruction
from long-term retention while keeping active conditioning bounded.
Figure~\ref{fig:method} illustrates our framework.

\subsection{Persistent Autoregressive Generation}
\label{sec:task}

\methodname{} generates video in consecutive segments from a single source
image and target camera poses. The source image $\mathbf I_0^{\mathrm{init}}$ forms the
initial Scenelet $\scenelet_0$, giving
$\historybank_0=\{\scenelet_0\}$.
For segment $k>0$, let $\mathbf I_k^{\mathrm{init}}$ denote its starting
image, $\queryset_k$ the target camera poses, and
$\historybank_k=\{\scenelet_0,\ldots,\scenelet_{k-1}\}$ the set of Scenelets
accumulated so far. Projecting the accumulated Scenelets into the target views
$\queryset_k$ and resolving visibility produces the bounded, target-aligned
condition $\mathbf C_k$. We augment the pretrained SANA-WM~\cite{zhu2026sanawm} backbone
$\theta$ with a learned ControlNet-style~\cite{zhang2023controlnet}
spatial-conditioning branch $\phi$. Let $\widehat{\mathbf V}_k$ denote the
generated segment and
$\scenelet_{k-1}$ its reconstructed Scenelet ($\scenelet_0$ from source image). The
memory-conditioned generator $\mathcal F_{\theta,\phi}$ operates causally:
\begin{equation}
\begin{gathered}
\widehat{\mathbf V}_k
= \mathcal F_{\theta,\phi}\bigl(\mathbf I_k^{\mathrm{init}},\queryset_k;\mathbf C_k\bigr),\\
\mathbf I_k^{\mathrm{init}}=\operatorname{last}(\widehat{\mathbf V}_{k-1}),
\qquad
\historybank_k
= \historybank_{k-1} \cup \{\scenelet_{k-1}\}.
\end{gathered}
\label{eq:causal-generation}
\end{equation}
The pretrained decoder produces $\widehat{\mathbf V}_k$ for autoregressive
feedback and Scenelet construction. Only $\mathbf I_0^{\mathrm{init}}$ and
$\scenelet_0$ come directly from the source observation. After initialization,
each segment starts from a generated image and appends a Scenelet built from
generated frames, so prediction errors can influence later segments.

\subsection{Camera Aware Ray Lifting}
\label{sec:camera-interface}

Persistent memory must retain the geometry and appearance captured across the
camera's full field of view, rather than being limited to the narrower
perspective view used for generation.
RecallWorld lifts each source pixel along its viewing ray using the corresponding
radial range, mapping it to a 3D point in the shared world space. Perspective and
fisheye images can therefore be represented in the same world space, including
surfaces outside the rectified perspective view.

For an observation $\mathbf I_i$, let $\mathcal C_i$ denote its calibrated
pinhole or Kannala--Brandt fisheye camera model~\cite{kannala2006generic} and
$\mathbf T_i = (\mathbf R_i,\mathbf t_i)$ its camera-to-world pose, where
$\mathbf R_i \in \mathrm{SO}(3)$ and $\mathbf t_i \in \mathbb R^3$.
Generated keyframes take their poses from the target trajectory. At image
coordinate $\mathbf x \in \mathbb R^2$, calibrated unprojection gives the unit
viewing ray $\mathbf r_i(\mathbf x)$, and radial range $\rho_i(\mathbf x)$
locates the corresponding world-space point:
\begin{equation}
\begin{gathered}
\mathbf r_i(\mathbf x)=\pi_{\mathcal C_i}^{-1}(\mathbf x),
\qquad \|\mathbf r_i(\mathbf x)\|_2=1,\\
\mathbf X_i(\mathbf x)
=\mathbf R_i[\rho_i(\mathbf x)\mathbf r_i(\mathbf x)]+\mathbf t_i.
\end{gathered}
\label{eq:calibrated-lift}
\end{equation}
Unlike optical depth, radial range remains defined for fisheye rays at or
beyond $90^\circ$ from the optical axis.

We introduce two \methodname{} variants that differ in source initialization
and initial memory coverage but share all subsequent processing, including the
use of generated perspective frames for later Scenelets.
\textbf{\navworldperspective{}} uses the perspective source image to initialize
both generation and $\scenelet_0$. Depth Anything~3~\cite{lin2025da3} estimates
its depth, which we convert to radial range. \textbf{\navworldfisheye{}} instead
initializes $\scenelet_0$ from a calibrated native fisheye image and generation
from its perspective resampling $\mathbf I_0^{\mathrm{init}}$.
UniK3D~\cite{piccinelli2025unik3d} estimates the native image's radial range
directly along calibrated rays. Figure~\ref{fig:camera} shows how
\navworldfisheye{} retains surfaces outside the initial perspective view and
uses them to condition a later revisit.

\input{figures/camera/figure}

\subsection{Episodic Scenelet Memory}
\label{sec:spatial-memory}
\label{sec:scenelet-consolidation}

Beyond locating observed surfaces in 3D, persistent memory depends on
consolidation granularity applied to memory, which determines how nearby views share a
reconstruction. Per-view reconstruction discards constraints among nearby
frames, whereas repeatedly reconstructing the growing history allows newly
generated content to revise earlier geometry. \methodname{} avoids both extremes
by grouping each segment into a bounded Scenelet whose views share a local
reconstruction that remains independent across episodes.

For each Scenelet $\scenelet_g$ with $g\geq1$, $\keyframes_g$ indexes five
evenly spaced keyframes from the corresponding completed segment.
$\pi^3\mathrm{X}$~\cite{wang2025pi3} jointly reconstructs the keyframes using their
intrinsics and target poses. A similarity transform then aligns the predicted
cameras to those poses, using Kabsch~\cite{kabsch1976rotation} for rotation and
the camera centers for scale and translation. The resulting depths are converted
to radial range and lifted (see Equation~\ref{eq:calibrated-lift}), ensuring that
stored geometry matches the generator's camera conditioning. For
each $i\in\keyframes_g$, $\scenelet_g$ stores episode-specific point map
$\mathbf X_i^{(g)}$, appearance features
$\mathbf F_i=\mathcal E(\mathbf I_i)$ from frozen video encoder $\mathcal E$,
and validity mask $W_i^{(g)}$ for valid non-sky geometry:
\begin{equation}
\scenelet_g=
\bigl\{(\mathbf X_i^{(g)},\mathbf F_i,W_i^{(g)})\mid i\in\keyframes_g\bigr\}.
\label{eq:scenelet-construction}
\end{equation}
Each tuple remains tied to view $i$, with $\mathbf X_i^{(g)}$ and $W_i^{(g)}$
specific to episode $g$. A boundary frame appearing in adjacent segments
therefore receives a separate record in each Scenelet. The source image forms
the single-view Scenelet $\scenelet_0$, and new Scenelets are appended without
revising existing geometry. 

\subsection{Target-Aligned Scenelet Retrieval}
\label{sec:retrieval}

To keep conditioning bounded as Scenelet memory grows with the
trajectory~\cite{yi2026worldkv,yang2026neoverse}, \methodname{} retains every
Scenelet but activates at most $K$ for each target segment. Let $\Omega$ be the
set of spatiotemporal locations $p$ in the target segment's latent grid; we omit
the segment index $k$ for clarity. For each candidate
$\scenelet_g\in\historybank_k$, projection into the target poses produces
\begin{equation}
\Pi(\scenelet_g,\queryset_k)=(\mathbf A_g,D_g,B_g).
\label{eq:projection}
\end{equation}
At each $p$, the nearest valid surface among the Scenelet's keyframes determines
projected appearance $\mathbf A_g(p)$ and target-camera depth $D_g(p)$. Binary
support $B_g(p)\in\{0,1\}$ indicates whether such a surface exists.

Ranking Scenelets individually by pose distance, recency, or view
overlap~\cite{xiao2026worldmem,yu2025contextmemory} can select redundant
candidates covering the same target region. Following~\cite{wang2026anchorweave,yu2025contextmemory}, \methodname{} instead scores each
candidate by the additional target locations it covers beyond the current
selection. Retrieval begins with selected index set $\selectedset=\varnothing$
and uncovered set $\mathcal U=\Omega$. At each step, it selects
\begin{equation}
g^\star
=\underset{\substack{\scenelet_g\in\historybank_k\\g\notin\selectedset}}{\arg\max}
\sum_{p\in\mathcal U}B_g(p).
\label{eq:coverage}
\end{equation}
After selecting $g^\star$, we add it to $\selectedset$ and remove each location
it supports from $\mathcal U$. The procedure stops after $K$ Scenelets or when
no candidate adds coverage. Because set coverage is monotone submodular, greedy
selection achieves at least $1-1/e$ of the optimal coverage with $K$
Scenelets~\cite{nemhauser1978analysis}. Earlier Scenelets can therefore be
selected when they cover regions missed by later ones.

\subsection{Visibility-Resolved Conditioning}
\label{sec:scenelet-conditioning}
\label{sec:visibility-conditioning}

Coverage-based retrieval determines which Scenelets support a target location,
but not which projected surface is visible. Independently reconstructed
Scenelets may project different surfaces onto the same target location, such as
a wall and the building behind it. Averaging their appearance features would
introduce occluded content, so \methodname{} resolves visibility geometrically
before fusion.

At a supported location $p$, let $D_{\min}(p)>0$ denote the nearest projected
depth among selected Scenelets with $B_g(p)=1$. Given a relative depth
tolerance $\delta\geq0$, define the visibility-compatible set
\begingroup
\thinmuskip=1mu \medmuskip=1mu \thickmuskip=1mu
\begin{equation}
\mathcal V(p)
=\bigl\{g\in\selectedset:\ B_g(p)=1,\ D_g(p)\le(1+\delta)\,D_{\min}(p)\bigr\}.
\label{eq:visible-scenelets}
\end{equation}
\endgroup
For each $g\in\mathcal V(p)$, binary weight $w_g(p)$ indicates whether its
projected appearance passes validity checks. If at least one
appearance is valid, the fused condition is
\begin{equation}
\mathbf C(p)=
\frac{\sum_{g\in\mathcal V(p)}w_g(p)\,\mathbf A_g(p)}
     {\sum_{g\in\mathcal V(p)}w_g(p)}.
\label{eq:spatial-condition}
\end{equation}
Depth-compatible projections are treated as the same surface, and valid
observations receive equal weight. A binary mask $M(p)$ indicates whether any
valid projected appearance remains, and $\mathbf C(p)=\mathbf 0$ wherever
$M(p)=0$. Because visibility depends only on geometry, a front
surface without valid appearance still occludes farther surfaces, leaving its
location unconditioned rather than exposing hidden content. Across the target
grid, $\mathbf C$ and $M$ form the segment condition $\mathbf C_k$ for the
spatial branch. Figure~\ref{fig:visibility-conditioning} illustrates this
geometry-first rule and its effect on the generated revisit.

\input{figures/visibility/figure}

\input{tables/02_experiment_floats}

\subsection{Training and Inference}
\label{sec:training-inference}

\noindent\textbf{Memory-conditioned training.}
The spatial branch copies a subset of backbone blocks, embeds $\mathbf C_k$,
and injects its outputs into the backbone through zero-initialized projections.
During this stage, only $\phi$ is trained; the backbone, video encoder, and
pretrained decoder remain fixed. We train the branch with the conditional
flow-matching objective used by
SANA-WM~\cite{lipman2022flow,zhu2026sanawm}.
For a recorded training segment with latent $\mathbf z=\mathcal E(\mathbf V)$
and noise $\boldsymbol\epsilon\sim\mathcal N(\mathbf 0,\mathbf I)$, the noisy
latent and target velocity are
\begin{equation}
\mathbf z_{\boldsymbol\tau}
=(\mathbf 1-\boldsymbol\tau)\odot\mathbf z+\boldsymbol\tau\odot\boldsymbol\epsilon,
\qquad
\mathbf u=\boldsymbol\epsilon-\mathbf z,
\label{eq:flow-path}
\end{equation}
where $\boldsymbol\tau$ holds one noise level per latent frame and $\odot$ is
elementwise multiplication. The first latent frame encodes the starting image
and remains clean ($\tau_0=0$); the remaining frames share one sampled noise
level in $[0,1]$. Let $m$ be a binary mask over the latent tensor that excludes
the first frame. The branch minimizes
\begin{equation}
\setlength{\abovedisplayskip}{2pt}
\setlength{\abovedisplayshortskip}{2pt}
\setlength{\belowdisplayskip}{2pt}
\setlength{\belowdisplayshortskip}{2pt}
\lossname_{\mathrm{FM}}(\phi)=\mathbb E\Bigl[
\tfrac{1}{\|m\|_1}\bigl\|m\odot
(\mathbf v_{\theta,\phi}-\mathbf u)\bigr\|_2^2\Bigr],
\label{eq:flow-matching}
\end{equation}
where $\mathbf v_{\theta,\phi}$ is the predicted velocity at
$(\mathbf z_{\boldsymbol\tau},\boldsymbol\tau;\allowbreak
\mathbf I_k^{\mathrm{init}},\queryset_k,\mathbf C_k)$, and $\|m\|_1$
counts the supervised latent entries. During training, $\mathbf C_k$ is formed
from preceding recorded frames. At inference, it is formed from the source
observation and prior model predictions.

\noindent\textbf{Decoder refinement.}
The pretrained decoder loses fine scene detail that the latents retain.
Holding the encoder and generator fixed, we fine-tune a copy of the decoder,
$\mathcal D_\psi$, for higher-fidelity reconstruction. For a recorded video
$\mathbf V$ with $L$ frames and $N$ RGB values, let
$\widetilde{\mathbf V}=\mathcal D_\psi(\mathcal E(\mathbf V))$ denote its
reconstruction. The decoder objective is
\begingroup
\thinmuskip=1mu \medmuskip=1mu \thickmuskip=2mu
\begin{equation}
\setlength{\abovedisplayskip}{4pt}
\setlength{\abovedisplayshortskip}{4pt}
\setlength{\belowdisplayskip}{5pt}
\setlength{\belowdisplayshortskip}{5pt}
\lossname_{\mathrm{dec}}(\psi)
=\tfrac{1}{N}\bigl\|\widetilde{\mathbf V}-\mathbf V\bigr\|_1
+\tfrac{1}{L}\textstyle\sum_{\ell=1}^{L}
\lossname_{\mathrm{LPIPS}}\bigl(\widetilde{\mathbf V}_\ell,\mathbf V_\ell\bigr),
\label{eq:decoder-refinement}
\end{equation}
\endgroup%
where $\ell$ indexes frames and $\lossname_{\mathrm{LPIPS}}$ is the LPIPS
perceptual loss~\cite{zhang2018lpips}. Table~\ref{tab:decoder-alignment}
shows that refinement improves both reconstruction and generation
fidelity on the same latents.

\noindent\textbf{Autoregressive inference.}
\methodname{} retrieves up to $K$ Scenelets, resolves visibility, and generates
each segment according to the causal update in
Equation~\ref{eq:causal-generation}. The pretrained decoder supplies the next
starting image and the
keyframes used for Scenelet construction. The refined decoder produces only the
reported output video.

%% file: figures/camera/figure.tex
\definecolor{navworldrect}{HTML}{1F6FD1}
\begin{figure}[t]
  \centering
  \def\navfigurewidth{\linewidth}
  \input{figures/camera/layout}
  \caption{\textbf{Wide-field memory on GrandTour~\cite{tuna2026grandtour}.}
  Region $R$ lies outside the rectified view, marked by the
  \textcolor{navworldrect}{dashed box} in the initial observation.
  From the rectified view, SANA-WM~\cite{zhu2026sanawm} has no memory of $R$ while \navworldperspective{} recalls it.
  \navworldfisheye{} initializes memory from the full fisheye view and retains
  $R$ sharply.}
  \label{fig:camera}
\end{figure}
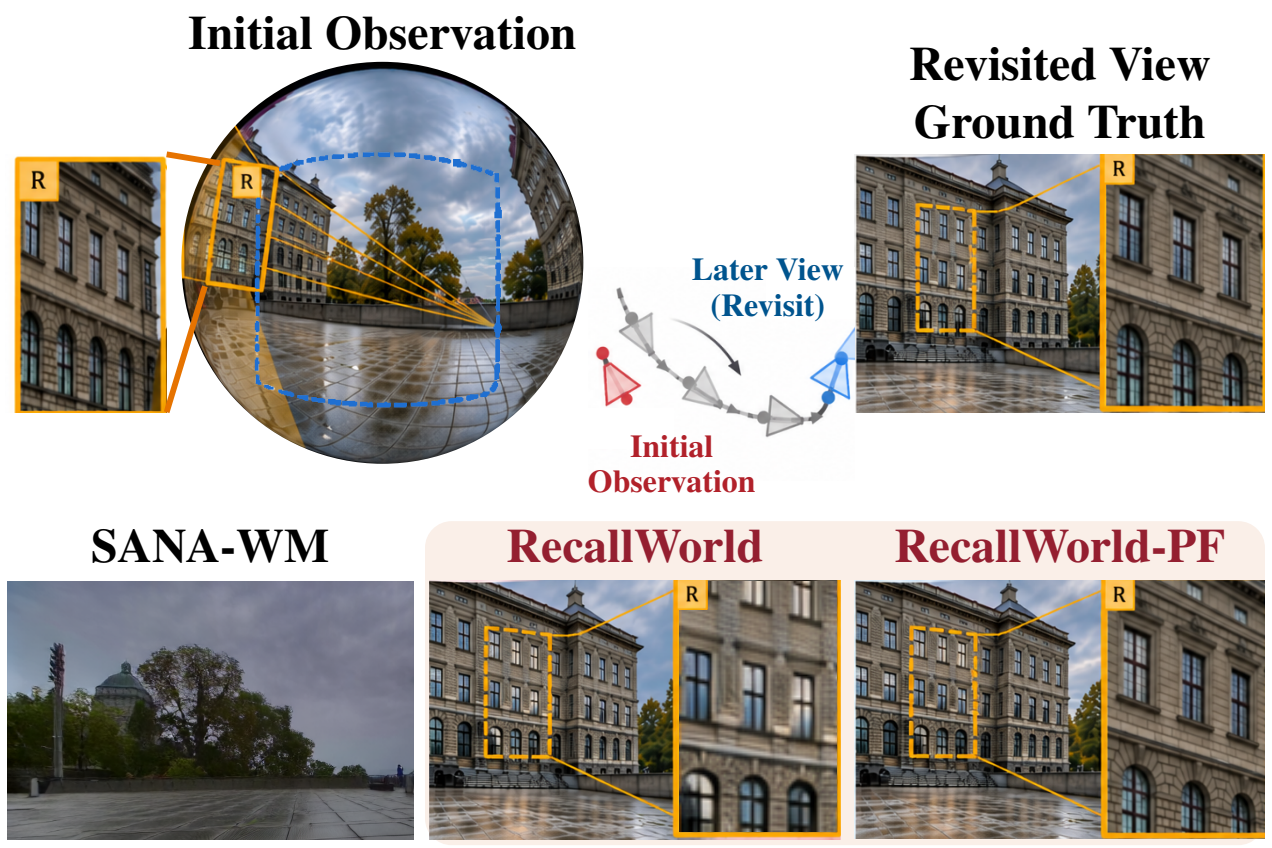

%% file: figures/camera/layout.tex
\begingroup%
\definecolor{cameraPlaceholder}{RGB}{148,163,184}%
\definecolor{initred}{RGB}{200,40,45}%
\definecolor{laterblue}{RGB}{30,110,200}%
\definecolor{pathgray}{RGB}{120,120,120}%
\begin{tikzpicture}[x=\navfigurewidth/1536,y=-\navfigurewidth/1536]
  \path[use as bounding box] (0,0) rectangle (1536,1045);
  \tikzset{cameraheading/.style={anchor=center,inner sep=0pt,
    font=\rmfamily\small\bfseries},
    pathlabel/.style={anchor=center,align=center,inner sep=0pt,font=\rmfamily\scriptsize}}

  \fill[navworldrow,rounded corners=5pt] (512,650) rectangle (1536,1045);

  \node[cameraheading] at (460,55) {Initial Observation};
  \node[cameraheading,align=center,anchor=south] at (1286,185) {Revisited View\\Ground Truth};
  \begin{scope}
    \clip (0.0,202.8) rectangle (197.0,517.2);
    \node[anchor=north west,inner sep=0pt] at (-13.05,188.51)
      {\includegraphics[width=1.1868\navfigurewidth]{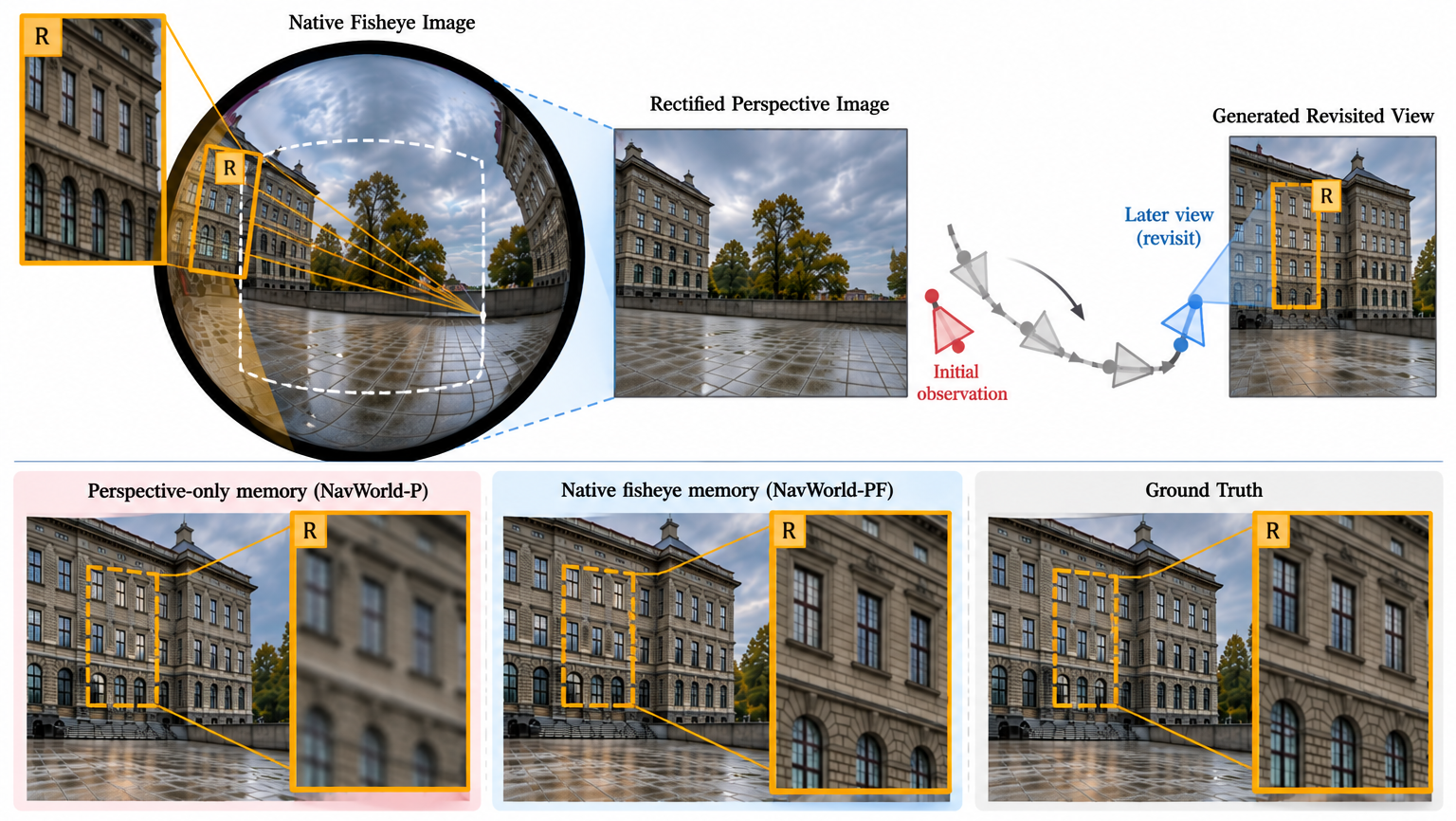}};
  \end{scope}
  \begin{scope}
    \clip (460.0,335.0) circle (245.0);
    \node[anchor=north west,inner sep=0pt] at (15.79,32.75)
      {\includegraphics[width=1.1449023\navfigurewidth]{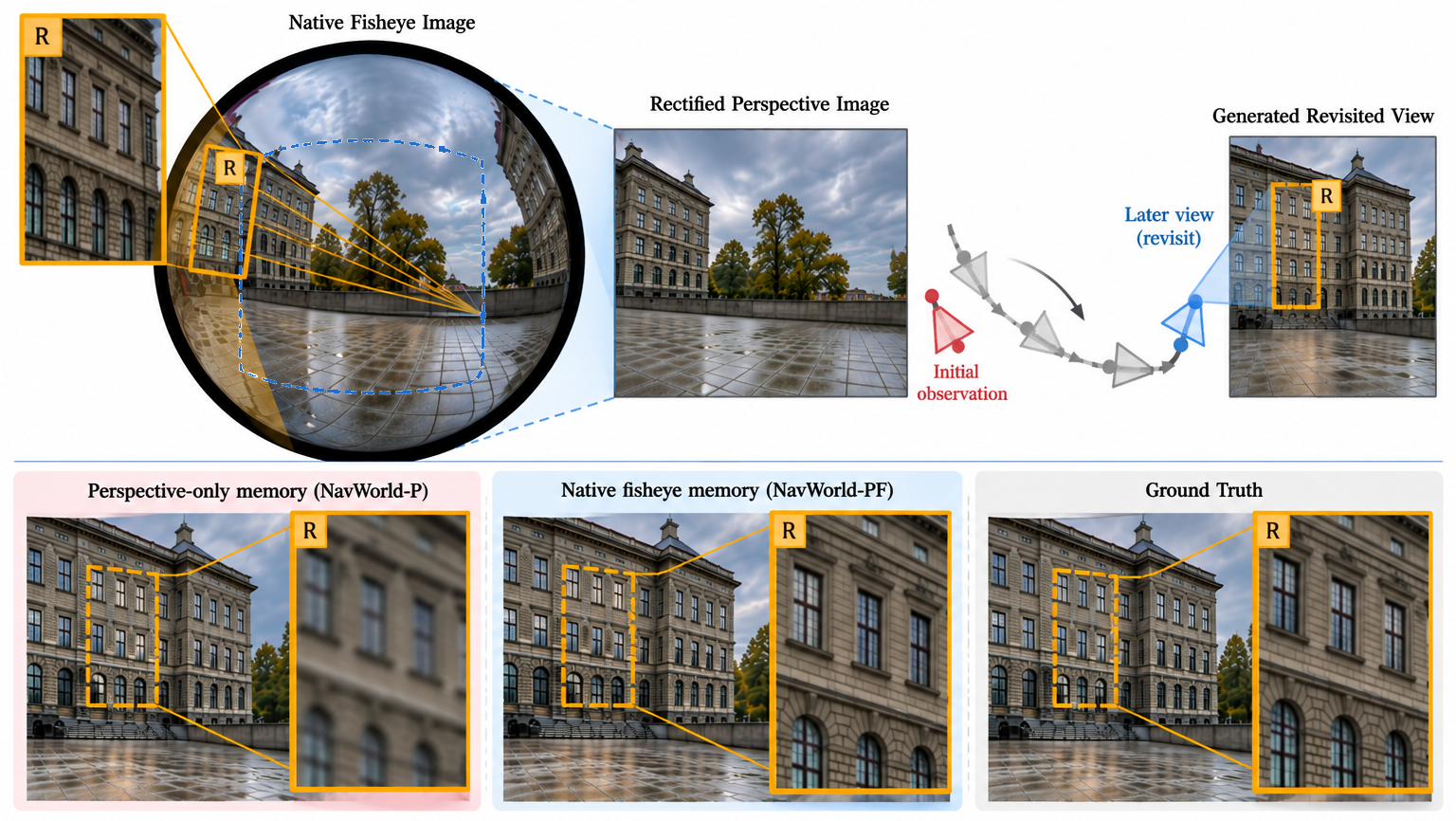}};
    \node[anchor=north west,inner sep=0pt] at (15.79,32.75)
      {\includegraphics[width=1.1449023\navfigurewidth]{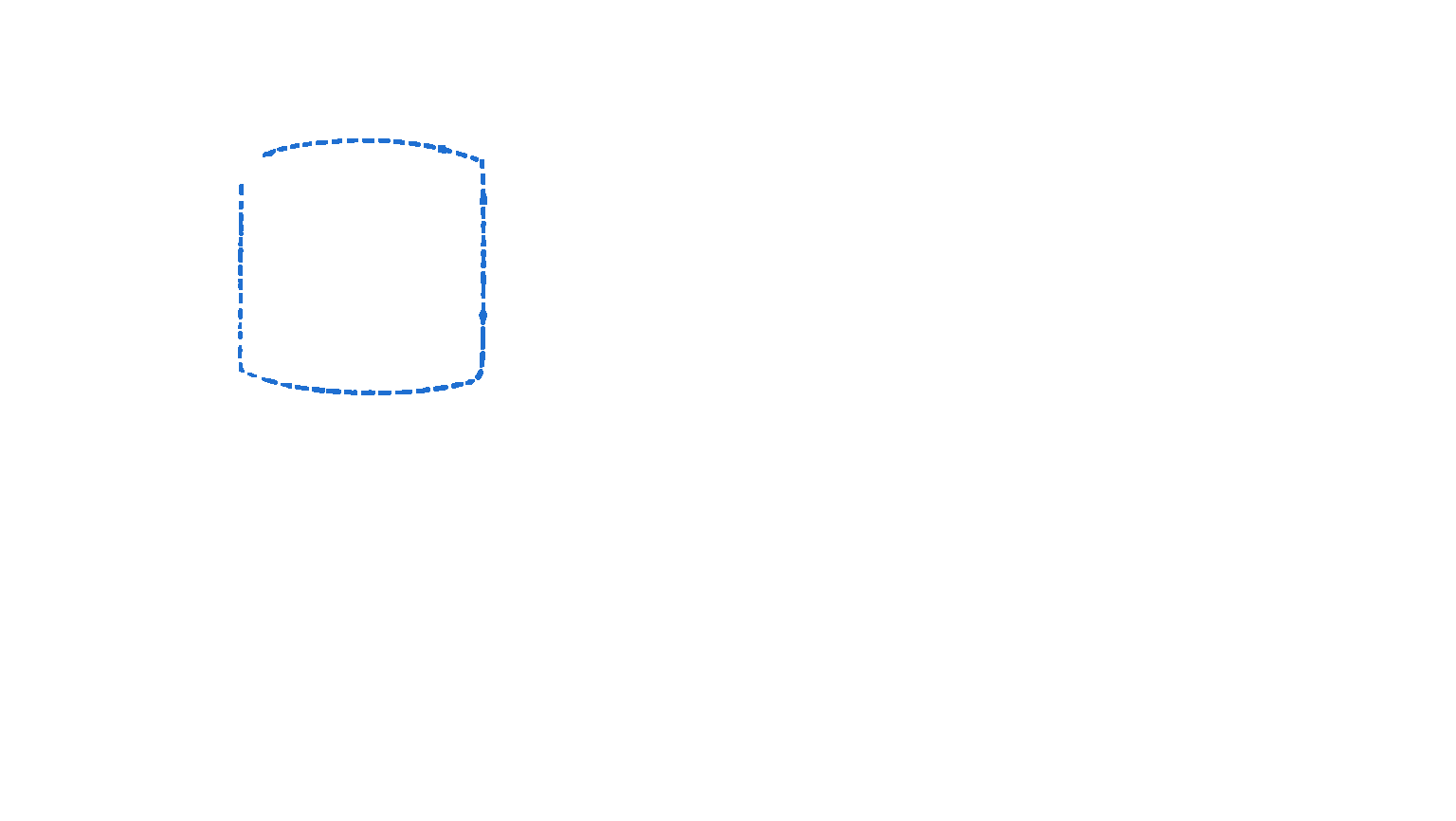}};
  \end{scope}
  \draw[orange!90!black,line width=1.1pt] (197.0,202.8) -- (261.93,213.61);
  \draw[orange!90!black,line width=1.1pt] (197.0,517.2) -- (243.63,363.64);
  \begin{scope}
    \clip (1036.0,202.8) rectangle (1536.0,517.2);
    \node[anchor=north west,inner sep=0pt] at (-74.87,-378.27)
      {\includegraphics[width=1.0661\navfigurewidth]{figures/camera/artwork.png}};
  \end{scope}
  \begin{scope}
    \clip (710,340) rectangle (1035,600.4167);
    \node[anchor=north west,inner sep=0pt] at (710,340)
      {\includegraphics[width=.2115885417\navfigurewidth]{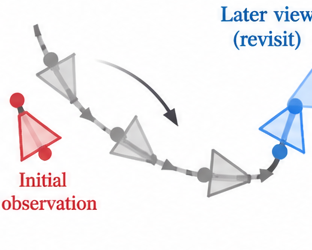}};
    \fill[white] (710,512.917) rectangle (815.208,561.875);
    \fill[white] (927.708,340) rectangle (1045.417,402.5);
  \end{scope}
  \node[anchor=center,align=center,inner sep=0pt,
    text={rgb,255:red,177;green,33;blue,42},
    font=\rmfamily\fontsize{6}{6.75}\selectfont\bfseries]
    at (812.778,580) {Initial\\Observation};
  \node[anchor=center,align=center,inner sep=0pt,
    text={rgb,255:red,0;green,84;blue,142},
    font=\rmfamily\fontsize{6}{6.75}\selectfont\bfseries]
    at (930,368.125) {Later View\\(Revisit)};

  \node[cameraheading] at (250,680) {SANA-WM};
  \node[cameraheading,text=navworldred] at (768,680) {\methodname{}};
  \node[cameraheading,text=navworldred] at (1286,680) {\methodname{}-PF};
  \begin{scope}
    \clip (2,722.0) rectangle (498,1038.5);
    \node[anchor=north west,inner sep=0pt] at (2,722)
      {\includegraphics[height=.2060546875\navfigurewidth]{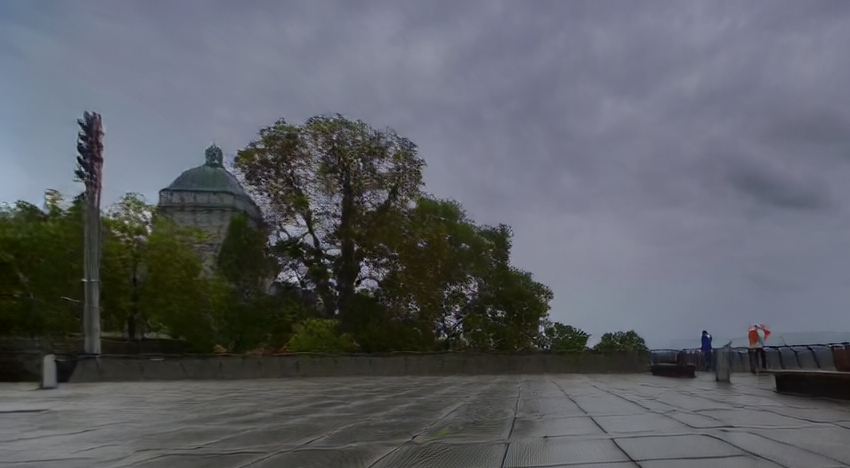}};
  \end{scope}
  \begin{scope}
    \clip (518.0,722.0) rectangle (1018.0,1038.5);
    \node[anchor=north west,inner sep=0pt] at (485.81,137.24)
      {\includegraphics[width=1.0730\navfigurewidth]{figures/camera/artwork.png}};
  \end{scope}
  \begin{scope}
    \clip (820.586,722.025) rectangle (1013.726,1027.830);
    \node[anchor=north west,inner sep=0pt] at (820.586,722.025)
      {\includegraphics[width=.1257421875\navfigurewidth]{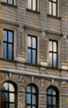}};
  \end{scope}
  \begin{scope}
    \clip (814.148,722.025) rectangle (855.995,757.434);
    \node[anchor=north west,inner sep=0pt] at (485.81,137.24)
      {\includegraphics[width=1.0730\navfigurewidth]{figures/camera/artwork.png}};
  \end{scope}
  \begin{scope}
    \clip (1036.0,722.0) rectangle (1536.0,1036.5);
    \node[anchor=north west,inner sep=0pt] at (469.90,140.98)
      {\includegraphics[width=1.0661\navfigurewidth]{figures/camera/artwork.png}};
  \end{scope}
\end{tikzpicture}%
\endgroup%

%% file: figures/visibility/figure.tex
\begin{figure}[t]
  \centering
  \captionsetup{skip=2pt}
  \def\navfigurewidth{\linewidth}
  \input{figures/visibility/layout}
  \par\vspace{-6pt}
  \caption{\textbf{Visibility resolution across Scenelets on
  GrandTour~\cite{tuna2026grandtour}.}
  (a) Independently reconstructed Scenelets project conflicting surfaces into
  the target view. (b) Nearest depth selects the frontmost surface.
  (c) Fusion without visibility resolution contaminates the generated revisit.
  (d) \methodname{} resolves the conflict before fusion and better matches
  (e) ground truth.}
  \label{fig:visibility-conditioning}
\end{figure}
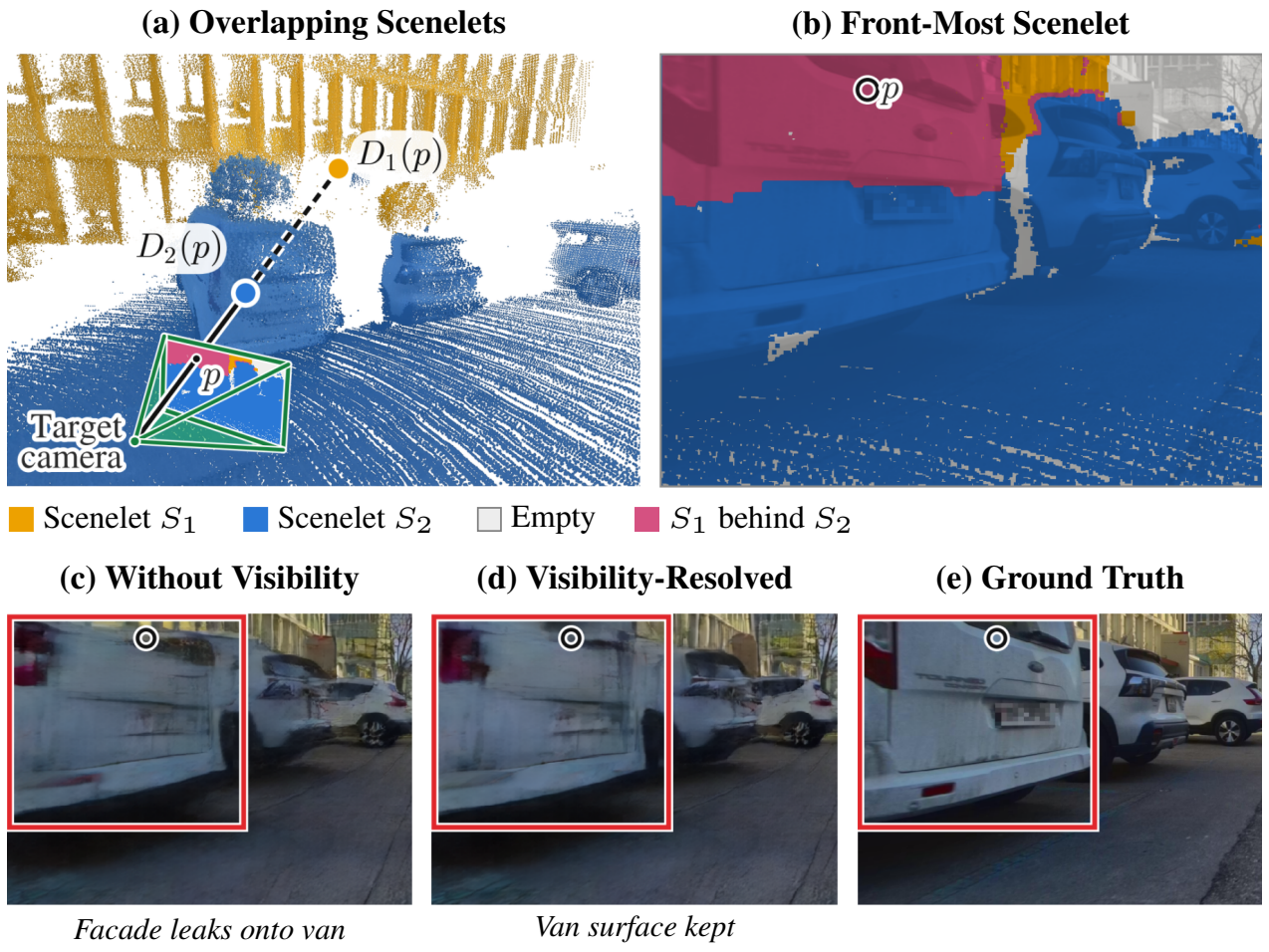

%% file: figures/visibility/layout.tex
\begingroup
\resizebox{\navfigurewidth}{!}{%
\begin{tikzpicture}[x=1bp,y=-1bp,every node/.style={inner sep=0pt,outer sep=0pt}]
\path[use as bounding box] (0,0) rectangle (255,192.685);
\tikzset{paneltitle/.style={font=\fontsize{6.8}{7.5}\selectfont\bfseries,anchor=center},
legendtext/.style={font=\fontsize{6.5}{7}\selectfont,anchor=west},
cropnote/.style={font=\fontsize{6.2}{7}\selectfont\itshape,anchor=center}}
\definecolor{viss1}{RGB}{237,161,0}
\definecolor{viss2}{RGB}{42,120,214}
\definecolor{visbehind}{RGB}{213,81,129}
\node[anchor=north west] at (0.0000,10.5025) {\includegraphics[width=128.8303bp]{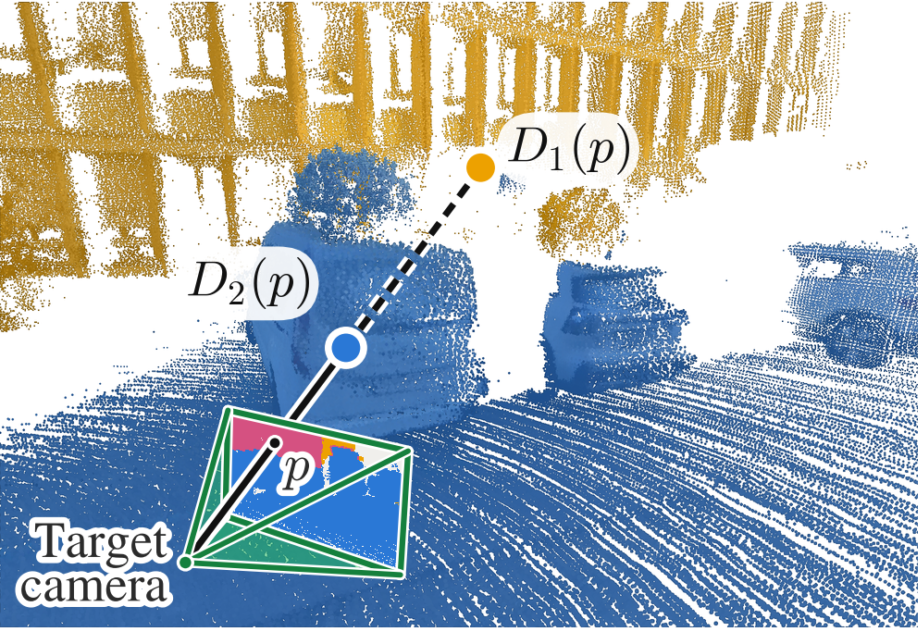}};
\node[anchor=north west] at (132.6112,10.5025) {\includegraphics[width=122.3888bp]{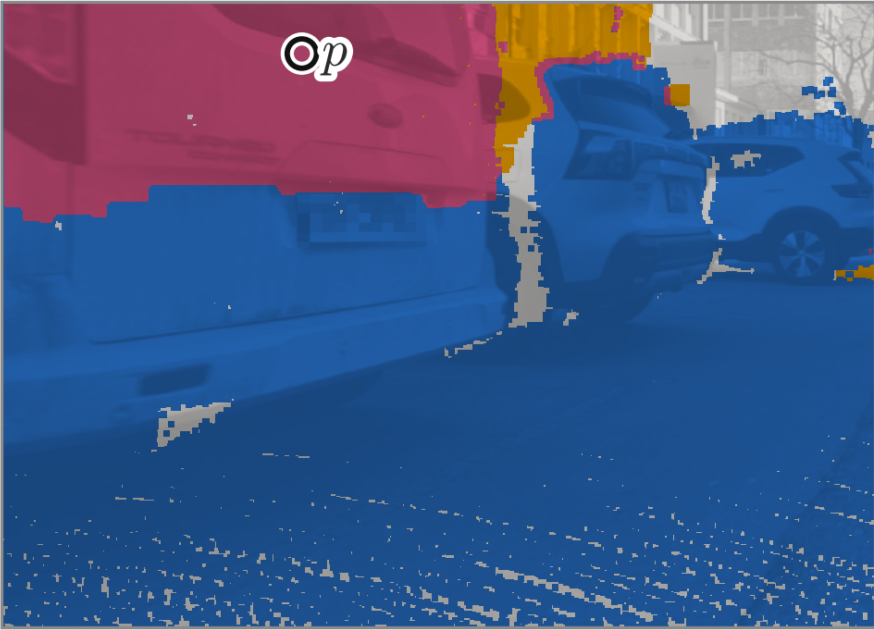}};
\node[anchor=north west] at (0.0000,124.3493) {\includegraphics[width=82.3394bp]{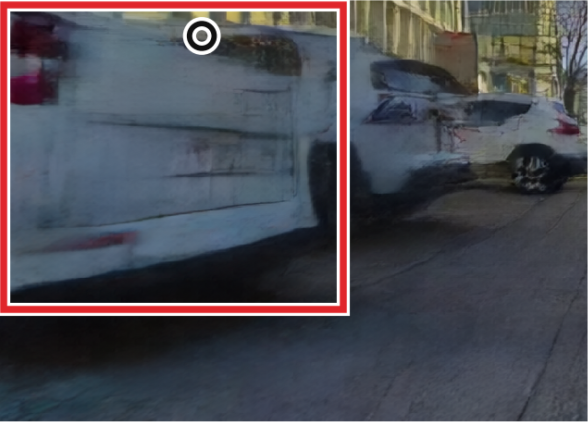}};
\node[anchor=north west] at (86.4003,124.3493) {\includegraphics[width=82.3394bp]{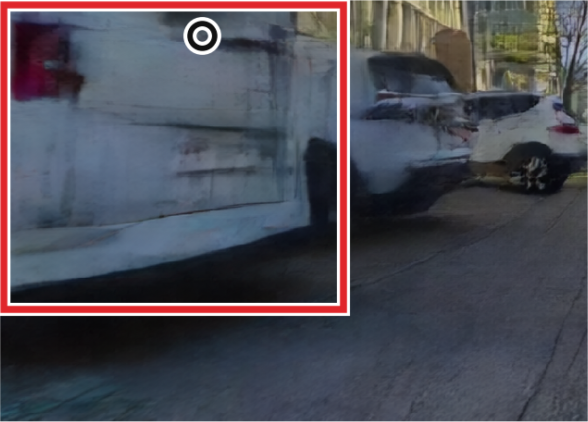}};
\node[anchor=north west] at (172.9407,124.3493) {\includegraphics[width=82.0593bp]{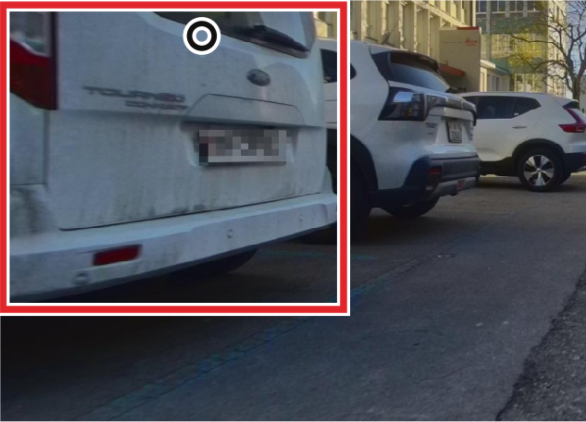}};
\node[paneltitle] at (64.417,4.65) {(a) Overlapping Scenelets};
\node[paneltitle] at (193.807,4.65) {(b) Front-Most Scenelet};
\node[paneltitle] at (41.17,117.5) {(c) Without Visibility};
\node[paneltitle] at (127.57,117.5) {(d) Visibility-Resolved};
\node[paneltitle] at (214.0,117.5) {(e) Ground Truth};
\path[fill=viss1,draw=viss1,line width=.35bp] (0.56,102.8) rectangle (5.26,107.5);
\node[legendtext] at (7.359999999999999,105.15) {Scenelet $S_1$};
\path[fill=viss2,draw=viss2,line width=.35bp] (48.18,102.8) rectangle (52.88,107.5);
\node[legendtext] at (54.98,105.15) {Scenelet $S_2$};
\path[fill=gray!14,draw=gray,line width=.35bp] (95.67,102.8) rectangle (100.37,107.5);
\node[legendtext] at (102.47,105.15) {Empty};
\path[fill=visbehind,draw=visbehind,line width=.35bp] (127.68,102.8) rectangle (132.38,107.5);
\node[legendtext] at (134.48000000000002,105.15) {$S_1$ behind $S_2$};
\node[cropnote] at (41.17,188.3) {Facade leaks onto van};
\node[cropnote] at (127.57,188.3) {Van surface kept};
\end{tikzpicture}%
}
\endgroup

%% file: tables/02_experiment_floats.tex
\setcounter{dbltopnumber}{1}
\input{tables/02_main_results}
\input{figures/revisits/figure}

%% file: tables/02_main_results.tex
\begin{table*}[!t]
\centering\small
\navworldtablestyle
\setlength{\tabcolsep}{1pt}
\caption{\textbf{Persistent embodied video generation.}
We compare camera-controlled and persistent-memory models on 32-second
exploration-and-return trajectories from GrandTour~\cite{tuna2026grandtour},
M2DGR~\cite{yin2021m2dgr}, TartanGround~\cite{patel2025tartanground}
and DL3DV~\cite{ling2023dl3dv}. Metrics cover revisit fidelity, camera
adherence and video consistency. Partial revisit averages high-,
moderate- and low-overlap cases. Best in \textbf{bold}, second best \underline{underlined}.}
\label{tab:main-results}
\label{tab:video-quality}
\renewcommand{\arraystretch}{1.06}
\begin{tabular*}{\textwidth}{@{\extracolsep{\fill}}l*{12}{c}@{}}
\toprule
 & \multicolumn{3}{c}{\textbf{Partial Revisit}}
 & \multicolumn{3}{c}{\textbf{Full Revisit (Cycle)}}
 & \multicolumn{3}{c}{\textbf{Camera Adherence}}
 & \multicolumn{3}{c}{\textbf{Video Consistency}}\\
\cmidrule(lr){2-4}\cmidrule(lr){5-7}\cmidrule(lr){8-10}\cmidrule(lr){11-13}
\textbf{Method}
 & PSNR\upmetric & SSIM\upmetric & LPIPS\downmetric
 & PSNR\upmetric & SSIM\upmetric & LPIPS\downmetric
 & TransErr\downmetric
 & RotErr\downmetric
 & ATE\downmetric
 & WS-3D\upmetric & VB-BG\upmetric & MEt3R\downmetric\\
\midrule
\rowcolor[HTML]{F3E7E9}
\multicolumn{13}{@{}l}{\textit{Camera-controlled generation}}\\
SANA-WM~\cite{zhu2026sanawm}
 & 12.82 & 0.366 & 0.710 & 13.96
 & 0.325 & 0.618
 & \underline{1.18\%} & \underline{1.43$^\circ$} & 3.00\% & 0.468 & 0.909 & 0.040\\
GEN3C~\cite{ren2025gen3c}
 & 13.22 & 0.320 & 0.732 & 14.06
 & 0.383 & 0.622
 & 2.01\% & 2.03$^\circ$ & 7.06\% & 0.628 & \underline{0.915} & 0.044\\
SEVA~\cite{zhou2025seva}
 & 13.67 & 0.377 & \underline{0.694} & 13.94
 & 0.350 & \underline{0.609}
 & \textbf{1.10\%} & \textbf{1.12$^\circ$} & \textbf{1.88\%} & \underline{0.635} & \textbf{0.930} & \textbf{0.028}\\
\midrule
\rowcolor[HTML]{F3E7E9}
\multicolumn{13}{@{}l}{\textit{Persistent-memory generation}}\\
VMem~\cite{li2025vmem}
 & 12.31 & 0.240 & 0.886 & 12.47
 & 0.249 & 0.841
 & 4.28\% & 8.89$^\circ$ & 11.94\%
 & 0.407 & 0.806 & 0.080\\
SPMem~\cite{wu2025spmem}
 & 13.74 & 0.313 & 0.782 & 13.31
 & 0.346 & 0.708
 & 4.00\% & 7.57$^\circ$ & 8.04\%
 & 0.448 & 0.853 & 0.073\\
Spatia~\cite{zhao2025spatia}
 & \underline{14.54} & \underline{0.379} & 0.728 & \underline{15.46}
 & \underline{0.402} & 0.616
 & 3.32\% & 6.61$^\circ$ & 8.15\%
 & 0.417 & 0.901 & 0.107\\
Lyra 2.0~\cite{shen2026lyra2}
 & 13.88 & 0.370 & 0.754 & 12.94
 & 0.359 & 0.731
 & 1.65\% & 2.60$^\circ$ & 6.34\%
 & 0.552 & 0.886 & 0.058\\
\midrule
\noalign{\begingroup
  \color{navworldrow}
  \dimen0=\dimexpr\ht\csname @arstrutbox\endcsname
    +\dp\csname @arstrutbox\endcsname\relax
  \hrule height\dimen0
  \vskip-\dimen0
\endgroup}
\methodname{}
 & \textbf{17.11} & \textbf{0.541}
 & \textbf{0.394} & \textbf{19.34}
 & \textbf{0.557} & \textbf{0.327}
 & 1.24\% & 2.43$^\circ$
 & \underline{2.81\%}
 & \textbf{0.685} & 0.914
 & \underline{0.035}\\
\bottomrule
\end{tabular*}
\end{table*}

%% file: figures/revisits/figure.tex
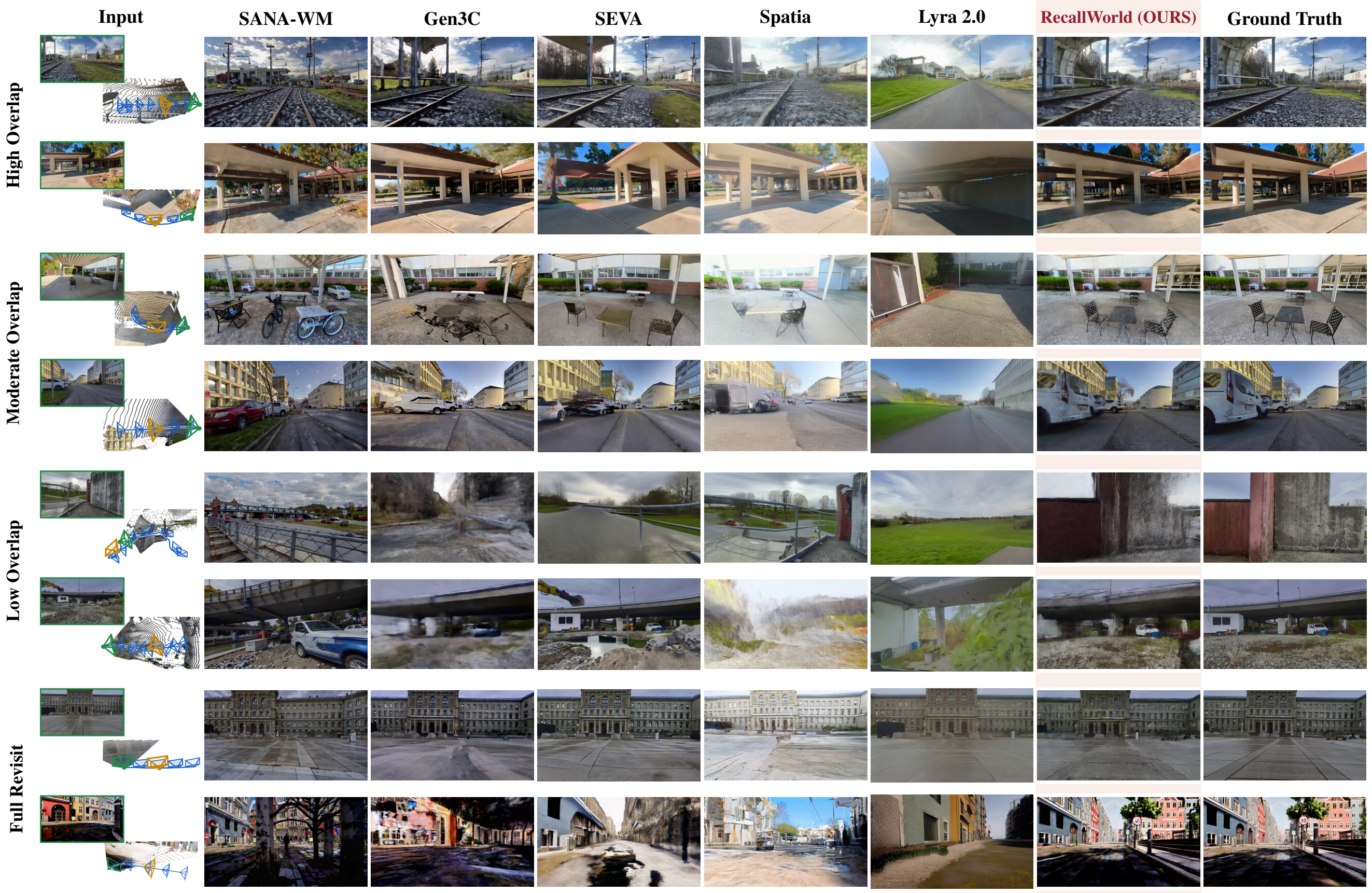
\begin{figure*}[t]
  \centering
  \resizebox{\textwidth}{!}{%
    \input{figures/revisits/layout.tex}}
  \captionsetup{skip=2pt}
  \caption{\textbf{Revisits on GrandTour~\cite{tuna2026grandtour},
  DL3DV~\cite{ling2023dl3dv} and TartanGround~\cite{patel2025tartanground}.}
  Rows are grouped into high-, moderate- and low-overlap partial revisits,
  followed by full retraced cycles. The input column shows the source image
  and camera path: \textcolor[HTML]{195FC9}{blue} marks the path,
  \textcolor[HTML]{228B45}{green} the source pose, and
  \textcolor[HTML]{C98A09}{gold} the revisit pose. The remaining columns compare
  generated revisits with recorded target views.}
  \label{fig:returns}
\end{figure*}

%% file: figures/revisits/layout.tex
\begin{tikzpicture}[x=1bp,y=-1bp,every node/.style={inner sep=0pt,outer sep=0pt}]
\path[use as bounding box] (0,0) rectangle (551.357,362.552);
\fill[white] (0,0) rectangle (551.357,362.552);
\fill[navworldrow,rounded corners=2bp] (417.588,0) rectangle (484.544,362.552);

\tikzset{figurefiveheading/.style={font=\rmfamily\fontsize{7.67}{8.2}\selectfont\bfseries}}
\node[figurefiveheading] at (47.5,7.8) {Input};
\node[figurefiveheading] at (114.3,7.8) {SANA-WM};
\node[figurefiveheading] at (181.6,7.8) {Gen3C};
\node[figurefiveheading] at (249.0,7.8) {SEVA};
\node[figurefiveheading] at (316.4,7.8) {Spatia};
\node[figurefiveheading] at (383.7,7.8) {Lyra 2.0};
\node[font=\rmfamily\fontsize{7}{7.6}\selectfont\bfseries,text=navworldred] at (451.1,7.8) {\methodname{} (OURS)};
\node[figurefiveheading] at (518.4,7.8) {Ground Truth};

\tikzset{figurefiverow/.style={rotate=90,font=\rmfamily\fontsize{7.37}{8}\selectfont\bfseries}}
\node[figurefiverow] at (4.5,55.1) {High Overlap};
\node[figurefiverow] at (4.5,143.2) {Moderate Overlap};
\node[figurefiverow] at (4.5,231.3) {Low Overlap};
\node[figurefiverow] at (5.2,319.5) {Full Revisit};

\def\figurefivecanvas#1{%
  \begin{tikzpicture}[x=1bp,y=1bp,every node/.style={inner sep=0pt,outer sep=0pt}]
    \path[use as bounding box] (-32.92,-19.60) rectangle (32.92,19.60);
    \fill[white] (-32.92,-19.60) rectangle (32.92,19.60);
    \node at (0,0) {#1};
  \end{tikzpicture}%
}
\def\figurefiveupright#1{%
  \figurefivecanvas{\includegraphics[width=65.84bp,height=39.2bp,keepaspectratio]{#1}}}
\def\figurefiveflipped#1{%
  \figurefivecanvas{\scalebox{1}[-1]{\includegraphics[width=65.84bp,height=39.2bp,keepaspectratio]{#1}}}}
\foreach \row/\y in {
  1/33.5,2/76.6,3/121.7,4/164.7,
  5/209.8,6/252.9,7/297.9,8/341.0}{
  \ifnum\row<7
    \let\figurefivepanel\figurefiveflipped
  \else
    \let\figurefivepanel\figurefiveupright
  \fi
  \node at (47.5,\y)
    {\figurefiveupright{figures/revisits/panels/r\row_input.png}};
  \node at (114.3,\y)
    {\figurefivepanel{figures/revisits/panels/r\row_sana.jpg}};
  \node at (181.6,\y)
    {\figurefivepanel{figures/revisits/panels/r\row_gen3c.jpg}};
  \node at (249.0,\y)
    {\figurefivepanel{figures/revisits/panels/r\row_seva.jpg}};
  \node at (316.4,\y)
    {\figurefiveupright{figures/revisits/panels/r\row_spatia.jpg}};
  \node at (383.7,\y)
    {\figurefiveupright{figures/revisits/panels/r\row_lyra.jpg}};
  \node at (451.1,\y)
    {\figurefivepanel{figures/revisits/panels/r\row_ours.jpg}};
  \node at (518.4,\y)
    {\figurefivepanel{figures/revisits/panels/r\row_gt.jpg}};
}
\end{tikzpicture}

%% file: sections/arxiv/04_experiments.tex
\section{Experiments}
\label{sec:experiments}
\label{sec:experimental-setup}

\looseness=-1
\methodname{} is based on the 
SANA-WM~\cite{zhu2026sanawm} backbone, generating 129-frame clips at
16\,fps. Five evenly spaced keyframes from each completed segment are jointly
reconstructed with $\pi^3\mathrm{X}$~\cite{wang2025pi3}. Retrieval
activates at most $K=4$ Scenelets. During memory-conditioned training, only the
copied spatial-conditioning blocks and their output projections are optimized;
the backbone, video encoder, and pretrained decoder remain fixed. The pretrained
decoder supplies autoregressive feedback and keyframes for Scenelet construction.
The encoder provides appearance features, and the refined decoder produces the
reported RGB outputs. We train on 8 H100 GPUs.

\noindent\textbf{Datasets and trajectories.}
We construct exploration-and-return trajectories from four datasets:
GrandTour~\cite{tuna2026grandtour} and M2DGR~\cite{yin2021m2dgr} provide real
robot navigation videos, TartanGround~\cite{patel2025tartanground} provides
synthetic navigation videos, and DL3DV~\cite{ling2023dl3dv} provides real scene
videos. The collection includes full revisits to earlier camera poses, following
loop-closure evaluations~\cite{li2025vmem,yang2026geometry}, and partial revisits
that reobserve previously seen regions from different viewpoints. Cycle
trajectories retrace the outgoing path in reverse; partial revisits span
\emph{high}, \emph{moderate}, and \emph{low overlap} with earlier views. The
main camera-pose comparison uses 32-second trajectories and common target poses.
Scores are averaged by clip.

\noindent\textbf{Baselines and configurations.}
Camera-controlled generation is represented by SANA-WM~\cite{zhu2026sanawm},
SEVA~\cite{zhou2025seva} and GEN3C~\cite{ren2025gen3c};
persistent-memory models include VMem~\cite{li2025vmem},
SPMem~\cite{wu2025spmem}, Spatia~\cite{zhao2025spatia}
and Lyra 2.0~\cite{shen2026lyra2}.
Table~\ref{tab:embodied-navigation} reports a separate comparison with
Navigation World Models (NWM)~\cite{bar2024nwm} on the 32-second navigation
subset. We map target camera poses to NWM's native planar-translation and yaw
actions, repeat the source image across its context slots, and score at 4\,Hz.
Unless stated otherwise, \methodname{} denotes the perspective variant.
External baselines retain their native encoders and decoders.

\setcounter{topnumber}{1}
\suppressfloats[t]
\input{tables/03_embodied_navigation}

\noindent\textbf{Metrics.}
\looseness=-1
Revisit fidelity uses PSNR~\cite{hore2010psnr},
SSIM~\cite{wang2004ssim}, and
LPIPS~\cite{zhang2018lpips} against recorded target views.
Following GEAR~\cite{yang2026geometry}, we align ViPE~\cite{huang2025vipe}
poses once per sequence with a similarity transform~\cite{umeyama1991alignment}
and report relative translation and rotation errors (TransErr and RotErr).
TransErr uses one-second intervals; ATE~\cite{sturm2012benchmark} measures
accumulated trajectory deviation. Both translation errors are normalized by
target path length and reported as percentages, while RotErr is reported in
degrees.
WorldScore 3D consistency~\cite{duan2025worldscore} and VBench background
consistency~\cite{huang2024vbench} assess spatial and temporal consistency.
MEt3R~\cite{asim2025met3r} measures cross-view feature disagreement after
3D alignment.
The navigation-command comparison additionally reports
DreamSim~\cite{fu2023dreamsim} and Scene Revisit Consistency
(SRC)~\cite{zhou2025persistentembodied}, which measures cosine similarity
between DINOv2~\cite{oquab2023dinov2} features from first-visit and return frames
at the same camera pose.

%% file: tables/03_embodied_navigation.tex
\begin{table}[t]
\centering\small
\navworldtablestyle
\setlength{\tabcolsep}{6pt}
\caption{\textbf{Navigation video generation.}
We compare RecallWorld with NWM~\cite{bar2024nwm} on the 32-second
navigation subset. NWM uses camera-derived movement actions.}
\label{tab:embodied-navigation}
\begin{tabular}{@{}lcccc@{}}
\toprule
\textbf{Method} & \textbf{PSNR\upmetric} & \textbf{LPIPS\downmetric}
 & \textbf{DreamSim\downmetric} & \textbf{SRC\upmetric}\\
\midrule
NWM~\cite{bar2024nwm}
 & 12.17 & 0.651 & 0.669 & 0.554\\
\rowcolor{navworldrow}
\methodname{} & \textbf{17.48} & \textbf{0.363} & \textbf{0.106} & \textbf{0.880}\\
\bottomrule
\end{tabular}
\end{table}

%% file: sections/arxiv/05_results.tex
\makeatletter
\let\recallworldnormalcvprsection\cvprsection
\def\cvprsection{\@startsection{section}{1}{\z@}
  {-4pt plus -1pt minus -1pt}{3pt}{\large\bf}}
\section{Results}
\let\cvprsection\recallworldnormalcvprsection
\makeatother
\label{sec:results}
\label{sec:baselines-metrics}
\label{sec:quality-experiments}

\looseness=1
\methodname{} achieves the best reported revisit fidelity across all three
metrics for both partial and full revisits in Table~\ref{tab:main-results}.
Against the strongest baseline for each metric, PSNR improves by 2.6\,dB
on partial revisits and 3.9\,dB on full cycles, while LPIPS decreases by
43.2\% and 46.3\%, respectively. \methodname{} also attains the highest
WorldScore 3D consistency and the second-lowest ATE. SEVA~\cite{zhou2025seva}
leads in camera adherence, background consistency and MEt3R but has
substantially worse revisit fidelity. Better trajectory adherence is therefore
not sufficient for faithful revisits. The fidelity advantage also carries over
to the separate navigation subset in Table~\ref{tab:embodied-navigation},
where \methodname{} outperforms NWM~\cite{bar2024nwm} on all four metrics,
raising SRC \mbox{from 0.554 to 0.880.}

\looseness=1
Figure~\ref{fig:returns} compares how faithfully each method recovers observed
scene content across partial and full revisits. SANA-WM~\cite{zhu2026sanawm}
generates a plausible street view but replaces the observed foreground van.
SEVA~\cite{zhou2025seva} likewise loses observed content despite its stronger
camera adherence. Spatia~\cite{zhao2025spatia} and Lyra 2.0~\cite{shen2026lyra2}
alter scene structure despite using persistent memory, showing that retaining
history does not by itself ensure faithful returns. By contrast, \methodname{}
more closely matches the recorded target views, preserving railway structures,
street layouts and building facades across both revisit types. Its advantage
is more faithful recovery of observed place identity, not visual plausibility
alone. 

\begin{table}[t]
  \captionsetup{skip=2pt}
  \input{tables/04_scenelet_memory}
\end{table}
\begin{figure}[t]
  \input{figures/persistent_access/figure}
\end{figure}
\input{figures/memory_ablation/figure}

\subsection{Memory and Decoder Ablations}
\label{sec:memory-results}

\looseness=-1
Local multi-view consolidation accounts for most of the PSNR gain in
Table~\ref{tab:memory-ablations}. Even without visibility resolution,
Scenelets improve PSNR over global memory by 3.7\,dB on partial revisits and
2.5\,dB on full cycles. Visibility resolution further improves both PSNR and
LPIPS. The complete model reduces LPIPS by 11.8\% and 7.1\%, respectively,
relative to independent-view memory. In Figure~\ref{fig:memory-qualitative},
global memory distorts scene structure and independent-view memory propagates
hallucinations. In Figure~\ref{fig:visibility-conditioning}, visibility resolution
prevents the occluded facade's appearance from leaking onto the foreground van.

\looseness=-1
Complementary retrieval achieves lower whole-view LPIPS than recent selection
at the longest history in Figure~\ref{fig:persistent-access}. Both use at most
$K=4$ active Scenelets on 64-second GrandTour~\cite{tuna2026grandtour} cycles.
The strategies coincide while stored memory fits within this budget. As history
grows, recent selection excludes earlier episodes and LPIPS rises, whereas
complementary retrieval recalls target-relevant Scenelets.
Scenelets lower peak memory versus global reconstruction by 19.7\%.

\begin{table}[t]
  \captionsetup{skip=2pt}
  \input{tables/06_decoder_alignment}
\end{table}

\looseness=-1
With latents held fixed, decoder refinement raises reconstruction PSNR by
6.1\,dB on GrandTour and 3.7\,dB on DL3DV~\cite{ling2023dl3dv} in
Table~\ref{tab:decoder-alignment}. Generation PSNR improves by 0.29\,dB and
0.35\,dB, respectively. GrandTour generation LPIPS
improves from 0.605 to 0.563. Generation fidelity drops by only $\sim1\%$ even without our fine-tuned decoder.
The pretrained decoder still supplies
autoregressive feedback and memory keyframes, leaving rollout history and
Scenelet construction unchanged.

\begin{figure}[t]
  \begingroup
    \captionsetup{type=table,hypcap=false,skip=2pt}
    \input{tables/05_camera_coverage}
  \endgroup
  \par\vspace{5pt}
  \input{figures/fisheye/figure}
\end{figure}

\subsection{Fisheye Memory}
\label{sec:camera-experiments}

\looseness=-1
Native fisheye memory retains observed surfaces lost to perspective
rectification. In Table~\ref{tab:camera-coverage}, \navworldfisheye{} increases
projected coverage from 52.67\% to 63.27\%, a 20.1\% relative increase,
and lowers both peripheral and whole-view LPIPS. With encoding and range
estimates held fixed, expanding the fisheye field of view from $90^\circ$ to
its native $120^\circ$~\cite{tuna2026grandtour} adds 2.58 percentage points
of coverage and improves whole-view LPIPS by 1.4\%.
Figures~\ref{fig:camera} and~\ref{fig:native-fisheye-generation} show sharper
revisited structure and coherent street layouts. Wide-field memory improves
perspective revisits without requiring fisheye generation.

%% file: tables/04_scenelet_memory.tex
\centering\small
\navworldtablestyle
\setlength{\tabcolsep}{1.3pt}
\caption{\textbf{Memory organization and visibility on GrandTour~\cite{tuna2026grandtour}.}
Revisit fidelity with memory reconstructed per keyframe, over the full history
or jointly per segment as Scenelets.
RecallWorld adds visibility resolution before Scenelet fusion.}
\label{tab:memory-ablations}
\renewcommand{\arraystretch}{1.04}
\begin{tabular*}{\linewidth}{@{\extracolsep{\fill}}lcccc@{}}
\toprule
\multirow{2}{*}[-5pt]{\shortstack[l]{\textbf{Memory}\\\textbf{Organization}}}
 & \multicolumn{2}{c}{\textbf{Partial Revisit}}
 & \multicolumn{2}{c}{\textbf{Full Revisit (Cycle)}}\\
\cmidrule(lr){2-3}\cmidrule(lr){4-5}
 & \makebox[0.142\linewidth]{PSNR\upmetric}
 & \makebox[0.142\linewidth]{LPIPS\downmetric}
 & \makebox[0.142\linewidth]{PSNR\upmetric}
 & \makebox[0.142\linewidth]{LPIPS\downmetric}\\
\midrule
Independent views
 & 16.41 & 0.365 & 16.27 & 0.255\\
Global memory
 & 17.02 & 0.344 & 17.80 & 0.241\\
Scenelets w/o visibility
 & 20.72 & 0.339
 & 20.27 & 0.238\\
\noalign{\begingroup
  \color{navworldrow}
  \dimen0=\dimexpr\ht\csname @arstrutbox\endcsname
    +\dp\csname @arstrutbox\endcsname\relax
  \hrule height\dimen0
  \vskip-\dimen0
\endgroup}
\methodname{}
 & \textbf{20.82} & \textbf{0.322}
 & \textbf{20.84} & \textbf{0.237}\\
\bottomrule
\end{tabular*}

%% file: figures/persistent_access/figure.tex
\begingroup
  \captionsetup{type=figure,hypcap=false,skip=3pt}
  \centering
  \includegraphics[width=0.90\linewidth]{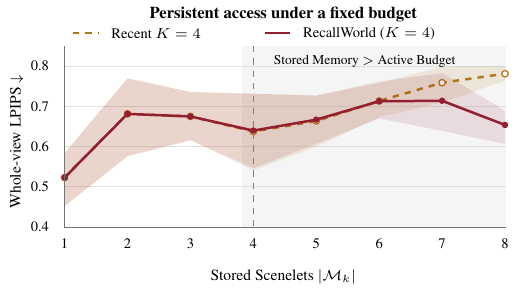}
  \caption{\textbf{Long-horizon fidelity with bounded Scenelet retrieval.}
  Whole-view LPIPS on 64 second GrandTour~\cite{tuna2026grandtour}
  cycles as stored memory grows under a $K=4$ Scenelet retrieval budget.}
  \label{fig:persistent-access}
\endgroup

%% file: figures/memory_ablation/figure.tex
\begin{figure*}[t]
  \centering
  \resizebox{\textwidth}{!}{%
    \input{figures/memory_ablation/layout.tex}}
  \captionsetup{skip=1.5pt}
  \caption{\textbf{Memory ablations on GrandTour~\cite{tuna2026grandtour}
  and TartanGround~\cite{patel2025tartanground}.}
  \methodname{} more closely matches the recorded target views, preserving local
  structure and detail better than global memory, independent-view memory, and
  Scenelets without visibility resolution.}
  \label{fig:memory-qualitative}
\end{figure*}

%% file: figures/memory_ablation/layout.tex
\begingroup%
\begin{tikzpicture}[x=1bp,y=-1bp,
  every node/.style={inner sep=0pt,outer sep=0pt}]
\path[use as bounding box] (0,0) rectangle (574,139);
\node[anchor=north west] at (0,0)
  {\includegraphics[width=574bp]{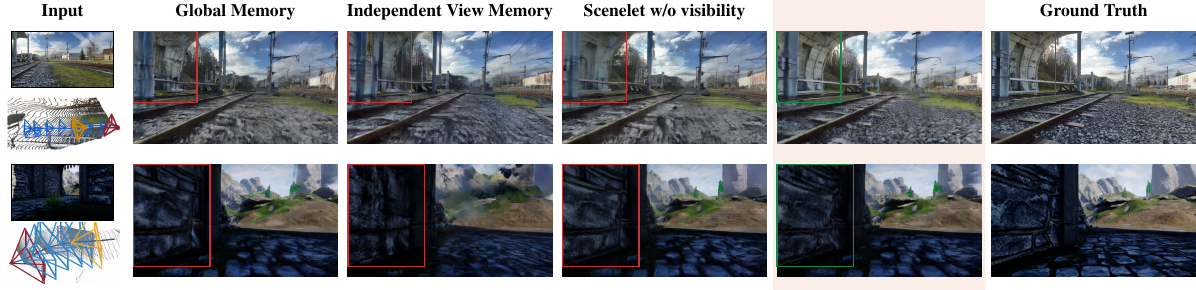}};
\node[font=\rmfamily\fontsize{8.5}{9.5}\selectfont\bfseries,
  text=navworldred] at (421.995,5.431)
  {\methodname{} (OURS)};
\end{tikzpicture}%
\endgroup%

%% file: tables/06_decoder_alignment.tex
\centering\small
\navworldtablestyle
\setlength{\tabcolsep}{1.3pt}
\caption{\textbf{Decoder refinement.}
Pretrained and refined decoders are applied to the same latents, from encoded
recorded videos for reconstruction and from whole RecallWorld rollouts for generation.}
\label{tab:decoder-alignment}
\renewcommand{\arraystretch}{1.0}
\begin{tabular*}{\linewidth}{@{\extracolsep{\fill}}llcccc@{}}
\toprule
 & & \multicolumn{2}{c}{\textbf{Reconstruction}}
 & \multicolumn{2}{c}{\textbf{Generation}}\\
\cmidrule(lr){3-4}\cmidrule(lr){5-6}
\textbf{Dataset} & \textbf{Decoder}
 & PSNR\upmetric & LPIPS\downmetric
 & PSNR\upmetric & LPIPS\downmetric\\
\midrule
\multirow{2}{*}{GrandTour~\cite{tuna2026grandtour}}
 & Pretrained
 & 19.16 & 0.301 & 13.93 & 0.605\\
 & \textcolor{navworldred}{\textbf{+ Refinement}}
 & \textbf{25.29} & \textbf{0.196}
 & \textbf{14.22} & \textbf{0.563}\\
\midrule
\multirow{2}{*}{DL3DV~\cite{ling2023dl3dv}}
 & Pretrained
 & 19.23 & 0.208 & 13.43 & 0.574\\
 & \textcolor{navworldred}{\textbf{+ Refinement}}
 & \textbf{22.92} & \textbf{0.188}
 & \textbf{13.78} & \textbf{0.568}\\
\bottomrule
\end{tabular*}

%% file: tables/05_camera_coverage.tex
\centering\small
\navworldtablestyle
\setlength{\tabcolsep}{1.3pt}
\caption{\textbf{Fisheye memory on GrandTour~\cite{tuna2026grandtour}.}
Coverage and revisit fidelity with perspective and native fisheye memory.
Peripheral scores cover regions outside the initial perspective view.}
\label{tab:camera-coverage}
\renewcommand{\arraystretch}{1.04}
\begin{tabular*}{\linewidth}{@{\extracolsep{\fill}}lccccc@{}}
\toprule
\multirow{2}{*}[-2pt]{\textbf{Method}}
 & \multicolumn{2}{c}{\textbf{Memory}}
 & \multicolumn{2}{c}{\textbf{Peripheral}}
 & \textbf{Whole}\\
\cmidrule(lr){2-3}\cmidrule(lr){4-5}\cmidrule(lr){6-6}
 & FoV & Coverage\upmetric
 & PSNR\upmetric & LPIPS\downmetric
 & LPIPS\downmetric\\
\midrule
\navworldperspective{} & $90^\circ$ & 52.67\% & 11.33 & 0.760 & 0.664\\
\navworldfisheye{} & $90^\circ$ & 60.69\% & 11.68 & 0.711 & 0.663\\
\noalign{\begingroup
  \color{navworldrow}
  \dimen0=\dimexpr\ht\csname @arstrutbox\endcsname
    +\dp\csname @arstrutbox\endcsname\relax
  \hrule height\dimen0
  \vskip-\dimen0
\endgroup}
\navworldfisheye{} & $120^\circ$ & \textbf{63.27\%} & \textbf{11.93} & \textbf{0.687} & \textbf{0.654}\\
\bottomrule
\end{tabular*}

%% file: figures/fisheye/figure.tex
\begingroup
  \captionsetup{type=figure,hypcap=false,skip=2pt}
  \centering
  \resizebox{\linewidth}{!}{%
    \input{figures/fisheye/layout.tex}}
  \caption{\looseness=-1\textbf{Generation from Fisheye image on GrandTour~\cite{tuna2026grandtour}.}
Existing world models~\cite{zhu2026sanawm} cannot account for fisheye
distortions, leading to severely distorted generations and blank periphery,
whereas \methodname{} retains structures beyond the initial $90^\circ$ FoV.}
  \label{fig:native-fisheye-generation}
\endgroup

%% file: sections/07_conclusion.tex
\section{Conclusion}
\label{sec:conclusion}

\looseness=-1
We introduced \textbf{\methodname{}}, a persistent embodied video world model
that separates local reconstruction from long-term retention. Episodic
Scenelets jointly reconstruct nearby views while keeping geometry independent
across episodes, so later predictions do not rewrite earlier reconstructions.
Complementary retrieval and geometry-first visibility resolution recall
relevant history within a fixed conditioning budget. Camera-aware ray lifting
unifies perspective and native fisheye observations in memory, retaining
wide-field content for later perspective generation. Across four benchmarks,
\methodname{} improves revisit PSNR over the strongest evaluated baseline by
2.6\,dB on partial revisits and 3.9\,dB on full cycles, demonstrating that
it more faithfully preserves place identity across exploration and return.

%% file: sections/06_limitations.tex
\par\noindent\textbf{Limitations and future work.}
\label{sec:limitations}
Long-horizon autoregressive generation can compound prediction errors,
motivating uncertainty-aware consolidation and cross-episode verification of
Scenelet memory. Extending static Scenelets with time-indexed geometry and
appearance could track moving actors, object changes, and illumination across
revisits while preserving enduring scene structure. Native fisheye generation
could further carry wide-field coverage from retained observations into
generated views. Together, these directions would extend persistent world
models from remembering where an agent has been to modeling how places evolve.

%% file: sections/arxiv/08_acknowledgements.tex
\section*{Acknowledgements}
This work was completed during an internship at Amazon.
We thank Ziyang Long, Amin Kheradmand, David Stafford, Yen-Liang Lin,
Shifan Zhu, Yun He, Guy Satat, Sravan Sripada, and Prakriti Chintalapoodi
for helpful discussions and feedback.